%% file: main.tex
\documentclass[runningheads]{llncs}

\usepackage{eccv}

\usepackage{eccvabbrv}
\usepackage[table]{xcolor}
\usepackage{graphicx}
\usepackage{booktabs}
\usepackage{enumitem}
\usepackage{colortbl}
\usepackage{amsmath,amssymb}
\usepackage{multirow}
\usepackage{tcolorbox}

\usepackage[accsupp]{axessibility}  % Improves PDF readability for those with disabilities.

\usepackage{hyperref}

\usepackage{orcidlink}

\begin{document}

% % 壓縮數學公式
% \setlength{\abovedisplayskip}{4pt}
% \setlength{\belowdisplayskip}{4pt}

% \setlength{\textfloatsep}{10pt plus 1.0pt minus 2.0pt} % 圖片跟內文的距離
% \setlength{\floatsep}{8pt plus 1.0pt minus 2.0pt}

\title{ECoSim: Data Efficient Fine-Tuning for Controllable Traffic Simulation} 

% ECoSim: Data-Efficient Control Adaptation for Traffic Simulation

\titlerunning{Data Efficient Fine-Tuning for Controllable Traffic Simulation}

\author{
Yu-Hsiang Chen\inst{1,2}\textsuperscript{*} \and
Wei-Jer Chang\inst{2}\textsuperscript{*} \and
Yi-Ting Chen\inst{1} \and
Masayoshi Tomizuka\inst{2}
}

\authorrunning{Y.-H. Chen et al.}

\institute{
National Yang Ming Chiao Tung University, Taiwan \and
University of California, Berkeley, USA
}

\maketitle

{\let\thefootnote\relax\footnotetext{* Equal contribution.}}

\begin{center}
  \includegraphics[width=0.9\linewidth]{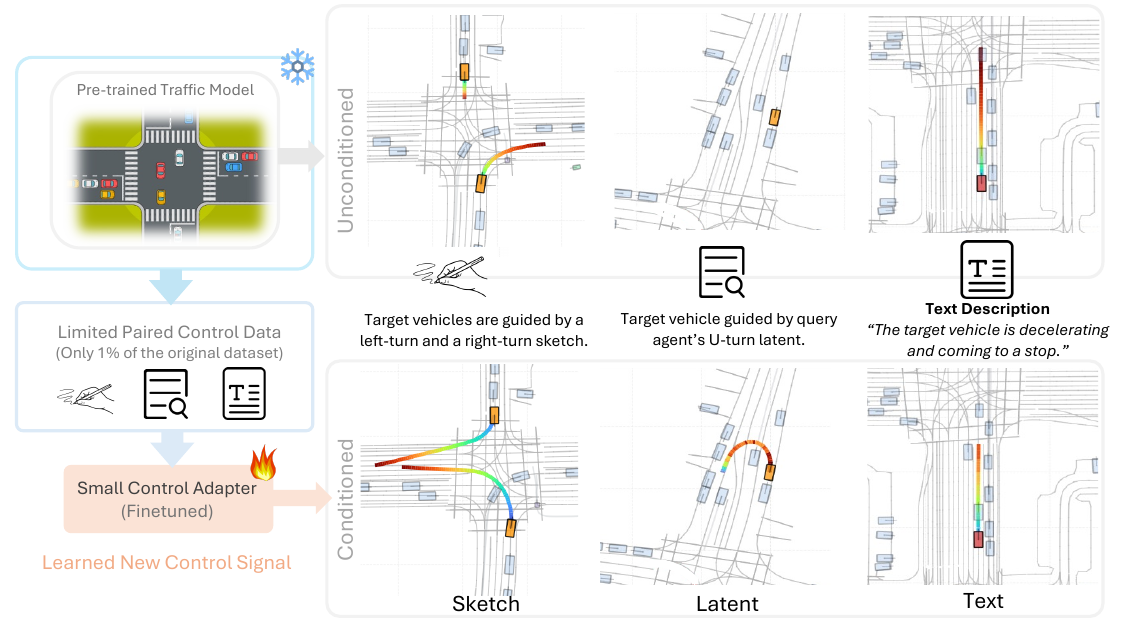}
  \captionof{figure}{\textbf{Data-Efficient Multi-Modal Control of Pretrained Traffic Models.} We introduce a data-efficient adaptation framework that steers frozen traffic models via sketch, latent, or text signals. Using only 1\% of the paired control data, the framework adds new control modalities and enables targeted generation of diverse user-specified traffic scenarios while preserving the pretrained model's performance.}
  \label{fig:teaser}
\end{center}

\input{sec/0_abstract}    
\input{sec/1_intro}
\input{sec/2_related_work}
\input{sec/3_method}

\input{sec/4_experiments}

\input{sec/5_disc}

\setcounter{secnumdepth}{0}

% \section{Acknowledgements}
% Please insert your acknowledgments here.

% \bibliography{main}

% % \clearpage
% % \appendix
% % \input{sec/6_suppl}

% \end{document}

\section{Acknowledgements}
This work was supported by the National Science and Technology Council under Grants 113-2628-E-A49-022 and 114-2628-E-A49-007, the H\&J Global Chair, the Ministry of Education, and the Yushan Fellow Program Administrative Support Grant. W.-J. Chang was supported by the National Science Foundation Graduate Research Fellowship Program under Grant DGE-2146752. The views expressed herein are those of the authors and do not necessarily reflect those of the National Science Foundation. Y.-H. Chen thanks the HCIS Lab at NYCU and the MSC Lab at UC Berkeley for their support.

% \section{Acknowledgements}
% This work was supported in part by the National Science and Technology Council under Grants 113-2628-E-A49-022 and 114-2628-E-A49-007, the H\&J Global Chair, the Ministry of Education, and the Yushan Fellow Program Administrative Support Grant. W.-J. Chang was also supported by the National Science Foundation Graduate Research Fellowship Program under Grant No. DGE-2146752. Any opinions, findings, conclusions, or recommendations expressed in this material are those of the authors and do not necessarily reflect the views of the National Science Foundation. Y.-H. Chen sincerely thanks the members of the HCIS Lab at National Yang Ming Chiao Tung University and the MSC Lab at UC Berkeley for their support.

\bibliographystyle{splncs04}
\bibliography{main}

% Uncomment below for supp.
\clearpage
\appendix
\setcounter{secnumdepth}{2}
\input{sec/6_suppl}

\end{document}

%% file: sec/0_abstract.tex
\begin{abstract}
Controllable traffic simulation is critical for testing autonomous driving systems, yet existing approaches often require retraining large generative models with extensive annotated data. We introduce a lightweight control adaptation framework that enables multi-modal controllability (sketch, latent behavior codes, and text) for pretrained state-of-the-art diffusion and autoregressive traffic models. By modulating intermediate features through identity-initialized FiLM layers, our method efficiently adds new control modalities while preserving the base model’s generative prior. Evaluated on Waymo Open Sim Agents Challenge, our approach demonstrates strong controllability with less than \textbf{1\%} of the paired control data. Through context-aware condition transfer, our framework enables counterfactual scenario generation and long-tail synthesis while maintaining stable closed-loop driving realism and safety. Our framework unlocks new possibilities for controllable traffic simulation, enabling targeted scenario generation through lightweight adaptation of pretrained generative models. Project page:
\href{https://ecosim-web.github.io/}
{\nolinkurl{https://ecosim-web.github.io/}}

% \keywords{Controllable Traffic Simulation \and Data-Efficient Adaptation}

\end{abstract}

% Generative traffic simulations often regress to nominal driving, severely limiting the synthesis of safety-critical and counterfactual scenarios. Existing controllable methods require costly full-parameter fine-tuning, while powerful autoregressive (AR) simulators remain notoriously difficult to guide. We introduce a lightweight control adaptation framework that endows frozen state-of-the-art simulators---both diffusion and AR backbones---with multi-modal controllability (sketch, latent, text). By modulating intermediate features via lightweight, identity-initialized FiLM layers, our approach achieves precise control without compromising the base generative prior. Evaluated on WOSAC, our adapter ($<$ 2.1M parameters) demonstrates extreme efficiency, achieving robust controllability with as little as $0.01\%$ of paired data and slashing training time by over $75\%$. Finally, through context-aware condition transfer, we synthesize diverse long-tail and counterfactual interactions while preserving the original model's performance in closed-loop driving realism, quality, and safety.

%% file: sec/1_intro.tex
\section{Introduction}
\label{sec:intro}
Traffic simulation is essential to autonomous vehicle (AV) development, enabling scalable and repeatable evaluation in controlled settings. It supports systematic validation under diverse driving conditions, shortens development cycles, and provides a safe platform for training and assessment. A key requirement is \textit{controllability}: the ability to deliberately induce targeted behavior patterns that probe specific system responses. Many traffic scenes are non-interactive and trivial, offering limited insight into failure modes. With controllability, simulation enables the construction of counterfactual ``what-if'' scenarios that expose potential weaknesses and improve diagnostic effectiveness.

Despite its importance, enabling controllable traffic simulation remains challenging. Prior approaches typically fall into two paradigms: inference-time guidance and direct conditioning through retraining. Inference-time methods, such as diffusion via optimization or gradient-based control during sampling~\cite{scenediffuser, vbd, safesim}, offer flexible guidance. However, such approaches often rely on carefully designed objectives and introduce additional computational overhead at inference. Alternatively, direct conditioning enables models~\cite{lctgen, prosim, langtraj, ctg} to internalize control signals during training, often yielding stable and faithful behavior. However, this comes at the cost of large-scale paired data and expensive retraining for each new control modality. Also, most existing evaluations focus on alignment with ground-truth trajectories, leaving the ability to generate controllable counterfactual scenarios across diverse contexts less explored.

%Our appraoch
% \ychen{this topic sentence should directly point out your contribution, instead of saying "we start ..." Use a sentence to summarize your work.}
%
% \yh{In this work, we introduce a data-efficient fine-tuning framework that endows pretrained traffic behavior model trained on large-scale, unconditional data with multi-modal controllability.}
% In this work, we start from a pretrained traffic behavior model trained on large-scale, unconditional data and adapt it through lightweight control adapters.
%
% \ychen{do you want to keep "parameter-efficient?"} \yh{update}
%

In this work, we introduce a data-efficient fine-tuning framework that enables controllable traffic simulation using traffic simulation models pretrained on large-scale data, hereafter referred to as traffic models. We freeze the base model and learn lightweight control adapters that inject conditioning signals into intermediate features, enabling efficient adaptation to new control modalities with limited paired control data. The adapters are trained under closed-loop simulation to enable controllability while preserving long-horizon interaction dynamics. Our framework supports multiple control modalities. Trajectory sketches provide precise spatial guidance, latent behavior embeddings extracted by our BehaviorVAE capture driving patterns learned from data, and natural language offers an intuitive user interface for specifying high-level behaviors. The same adaptation mechanism generalizes across different model paradigms, including diffusion-based and autoregressive traffic models. Finally, we introduce a context-aware retrieval pipeline that transfers learned behaviors into compatible scenes, enabling counterfactual and long-tail scenario generation.

We evaluate on the Waymo Open Sim Agents Challenge (WOSAC)~\cite{wosac} under closed-loop simulation. Across both diffusion-based and autoregressive models, our lightweight adapters reduce control error by up to 83\% while improving closed-loop realism by 0.02–0.03 absolute points in the WOSAC Meta score (see Sec.~\ref{subsec:main_results}). Remarkably, these gains are achieved using as little as 0.01\%–1\% paired fine-tuning data. Despite this data efficiency, our method matches or surpasses fully fine-tuned baselines trained on substantially larger datasets \cite{prosim}. 

Our main contributions can be summarized as follows:
\begin{itemize}[leftmargin=*, noitemsep, topsep=0pt, parsep=0pt]
    \item We propose a data-efficient control adaptation framework that adds multi-modal controllability (\emph{sketch}, \emph{latent}, and \emph{text}) to both diffusion and state-of-the-art autoregressive traffic models.
    
    \item We demonstrate strong \textbf{sample efficiency}. With identity-initialized FiLM adapters, our method achieves competitive controllability using minimal paired supervision (as few as 500 scenarios, 0.1\% of the paired control data) while keeping the pretrained backbone frozen.
    
    \item The proposed context-aware condition transfer mechanism enables counterfactual and long-tail scenario generation while preserving closed-loop realism.
    
    % We enable context-aware condition transfer for \textbf{counterfactual and long-tail scenario generation}. 
    % %
    % \ychen{this sentence sounds odd. Instead: The proposed context-aware condition transfer mechanism shows effective counterfactual and long-tail scenario generation.}
    % %
    % Our framework synthesizes diverse safety-critical multi-agent interactions while preserving closed-loop realism. 
    % \ychen{This contribution has not discussed in the introduction. How does it happen? } \yh{see above}
\end{itemize}

%% file: sec/2_related_work.tex
\section{Related Work}
\label{sec:related}

\begin{figure}[t]
  \centering
  \includegraphics[width=0.9\linewidth]{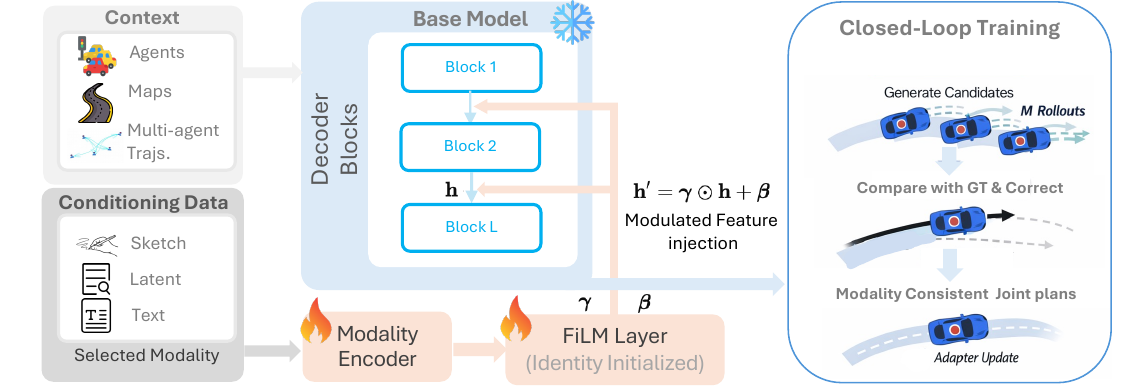}
  \caption{\textbf{Model-Agnostic Control Adaptation Architecture.} A lightweight adapter injects multi-modal control signals into a frozen pretrained traffic model by predicting FiLM parameters ($\boldsymbol{\gamma}, \boldsymbol{\beta}$) that modulate intermediate features $\mathbf{h}$. The design is compatible with both autoregressive and diffusion backbones, and identity initialization preserves the base model’s generative prior.}
  \label{fig:ctrl_architecture}
\end{figure}

% \noindent\textbf{Learning-based Traffic Simulation.} 
% To model multi-agent driving behaviors from real-world data~\cite{waymo}, early generative approaches (e.g., VAEs, GANs)~\cite{trafficsim_vae, social_gan} outperformed heuristic simulators~\cite{sumo} but struggled with long-horizon interactions. Recent advances overcome this using large Transformer-based architectures, predominantly leveraging diffusion~\cite{vbd, scenariodiffusion} and autoregressive (AR) or reactive~\cite{simnet, waymax, smart, behaviorgpt, catk} formulations. While diffusion models excel at capturing complex joint distributions for high-fidelity rollouts, AR simulators effectively model reactive behaviors by predicting motions step-by-step. Together, these paradigms have achieved state-of-the-art realism in benchmarks like WOSAC~\cite{wosac}. However, because these state-of-the-art models are predominantly trained unconditionally, their generated trajectories inherently regress to frequent, nominal driving patterns. This bias makes rare, safety-critical scenarios difficult to reproduce on demand, fundamentally limiting their utility for targeted stress-testing and counterfactual evaluation.

% MAIN LOGIC
%	1.	 Typical controllable generation appraoch(guidance, direct training,langtraj,prosim)
	% 2.	Autoregressive simulators (SOTA but not controllable)
	% 3.	How your method differs
	% 4.	Retrieval methods (RealGen) and how you differ

%Test-time guidance, direct training,
\noindent\textbf{Controllable Traffic Scenario Generation.} Recent work has made significant progress in building data-driven, realistic traffic models that support multi-agent interactions in closed-loop environments~\cite{waymax, simnet, symphony}. Recent autoregressive (AR) and reactive traffic models achieve strong closed-loop realism by modeling traffic generation as sequential next-token prediction~\cite{smart, catk, behaviorgpt}. However, these models are primarily designed for unconditional simulation and provide limited mechanisms for controllable scenario synthesis, which is crucial for structured, target-case testing~\cite{trafficgen, strive}.

Several works explore the generation of controllable traffic scenarios to synthesize targeted or safety-critical situations~\cite{trafficgen, safesim, chen2026collision}. Existing approaches can be broadly categorized into three mechanisms:
(1) \emph{Inference-time guidance} steers unconditional samples during generation via gradient-based or constraint-based objectives~\cite{ctd, scenediffuser}. While flexible, it often introduces significant inference overhead.
(2) \emph{Direct conditioning} trains generative models from scratch to internalize explicit control signals (e.g., goal specifications, scene constraints or language descriptions)~\cite{lctgen, prosim, langtraj, ctg}. However, introducing new modalities requires expensive retraining.
(3) \emph{Retrieval-based control} guides generation using similar scenarios retrieved from large datasets, such as RealGen~\cite{realgen}. 
In contrast, we adapt pretrained traffic models to support new control modalities without retraining the backbone. Our framework further extends retrieval-based guidance by introducing multiple control interfaces, including behavior latents learned via BehaviorVAE, enabling context-aware control within closed-loop simulation.

\noindent\textbf{Control Adaptation in Generative Models.}
Recent works in image and video synthesis explore adapting pretrained generative models to enable controllable generation. Rather than retraining large conditional models from scratch, these approaches augment pretrained backbones with mechanisms that incorporate new conditioning signals. 
For example, ControlNet~\cite{controlnet} introduces dedicated control branches connected via zero-initialized convolutions that progressively learn to modulate a pretrained diffusion model in response to structural inputs such as edges or poses.  Similarly, LoRA~\cite{lora} adapts pretrained models by injecting low-rank updates into existing weight matrices, enabling new capabilities while preserving the original model parameters. These works demonstrate that powerful generative models can be extended to new control modalities by leveraging pretrained generative priors.

Despite their success in visual generation, such control adaptation strategies remain largely unexplored in traffic modeling, where preventing compounding errors requires closed-loop training. We introduce a control adaptation framework optimized under closed-loop simulation to enable controllability while preserving the behavior of pretrained traffic models.

%% file: sec/3_method.tex
\section{Method}
\label{sec:method}

% We present a unified framework for controllable traffic simulation that steers a designated subset of agents via multiple modalities while maintaining realistic, closed-loop multi-agent interactions.
% We consider three distinct control signals: trajectory sketches, latent behavior codes, and natural language.
% To integrate these efficiently, we propose a \emph{Control Adaptation} mechanism that maps diverse inputs into a compact, shared embedding space and injects them into a frozen, pretrained base model via lightweight adapter layers.

\subsection{Problem Formulation}
 
Let $\pi_{\textrm{ref}}$ denote a model pre-trained on large-scale, unconditional driving data that generates future trajectories in a closed-loop manner given a scene context $c$.
Our goal is to extend this model to conditional generation, $\pi_{\theta}(\mathbf{S}_{t:t+T} \mid c, r)$, where $r$ denotes a control signal (e.g., trajectory sketches, latent behavior codes, or language descriptions) and to enable controllable multi-agent generation in a data-efficient manner.
The notation $\mathbf{S}_{t:t+T}$ denotes the joint multi-agent future states over a prediction horizon.
The generator should control designated agents according to $r$ while preserving realistic interactions of the pre-trained model $\pi_{\textrm{ref}}$.
For each control modality $m$, we assume access to a small control-annotated dataset $\mathcal{D}_m \subset \mathcal{D}$ and aim to adapt the pretrained traffic models with limited supervision.
%
% Our objective is to efficiently adapt the pretrained simulator using $\mathcal{D}_m$, such that controllability can be achieved with limited supervision. 
%
% To this end, we augment the pretrained model with an additional adaptation branch that incorporates the control signal $r$, while keeping the backbone parameters frozen to preserve its closed-loop behavior.

% To this end, we augment the pretrained model with a small set of additional trainable adaptation parameters \ychen{Do you want to keep this phrasing?} that incorporate the control signal $r$, while keeping the backbone parameters frozen to preserve its closed-loop behavior.

\subsection{Control Adaptation via Feature Modulation}

Given a raw control input $r_m$ from modality $m$, we first map it to a unified $d$-dimensional conditioning embedding $z \in \mathbb{R}^d$ via a modality-specific encoder (where $d=256$ in our implementation). For each agent $i$, only targeted agents receive their corresponding $z_i$, while non-targeted agents are assigned $z_i = \mathbf{0}$.

To enable controllable generation while preserving the pretrained traffic model prior, we inject the conditioning embedding $z_i$ into intermediate representations of the frozen backbone through lightweight feature modulation. We adopt Feature-wise Linear Modulation (FiLM)~\cite{film}, which conditions intermediate activations via an affine transformation. For an intermediate feature map $\mathbf{h}^{(l)} \in \mathbb{R}^{D}$ at the $l$-th decoder layer (where $D$ is the hidden dimension of the backbone), FiLM applies:
\begin{equation}
\mathrm{FiLM}(\mathbf{h}^{(l)}; z_i) 
= \boldsymbol{\gamma}^{(l)}(z_i) \odot \mathbf{h}^{(l)} 
+ \boldsymbol{\beta}^{(l)}(z_i),
\label{eq:film_multiply}
\end{equation}
where $\boldsymbol{\gamma}^{(l)}, \boldsymbol{\beta}^{(l)} \in \mathbb{R}^{D}$ are layer-specific modulation functions (implemented as lightweight MLPs) predicted from $z_i$.

A key design principle of our framework is extensibility: pretrained traffic models should be incrementally augmented with new controllability without retraining or destabilizing the original model. 
To this end, we initialize the modulation layers near identity ($\boldsymbol{\gamma}^{(l)} \approx \mathbf{1}$, $\boldsymbol{\beta}^{(l)} \approx \mathbf{0}$), ensuring that the adapted model initially approximates the base model.
This identity-preserving initialization is particularly important in closed-loop traffic simulation, where small perturbations can compound over long horizons and disrupt multi-agent interactions. 
Conceptually related to zero-initialized conditioning strategies such as ControlNet~\cite{controlnet}, this design enables controllability to be learned gradually through lightweight additional parameters while maintaining the backbone’s learned dynamics. 
% \ychen{what is the difference to controlnet?} \yh{controlnet only have one param, so it is zero init. }

% \subsection{Control Adaptation via Feature Modulation
% }

% We inject the encoded control signal $\mathbf{c}$ into the frozen base models across all $L$ layers of their respective backbones using Feature-wise Linear Modulation (FiLM)~\cite{film}. 
% For an intermediate feature map $\mathbf{h}^{(l)}$ at the $l$-th layer ($l \in \{1, \dots, L\}$), FiLM applies an affine transformation conditioned on $\mathbf{c}$:
% \begin{equation}
% \label{eq:film_affine}
% \mathrm{FiLM}(\mathbf{h}^{(l)}; \mathbf{c}) = \boldsymbol{\gamma}^{(l)}(\mathbf{c}) \odot \mathbf{h}^{(l)} + \boldsymbol{\beta}^{(l)}(\mathbf{c}),
% \end{equation}
% where $\boldsymbol{\gamma}^{(l)}$ and $\boldsymbol{\beta}^{(l)}$ are layer-specific modulation parameters predicted from $\mathbf{c}$. These are initialized near identity ($\boldsymbol{\gamma}^{(l)} \approx \mathbf{1}, \boldsymbol{\beta}^{(l)} \approx \mathbf{0}$) to strictly preserve the base model's generative prior at the start of training.

\subsection{Modality-Specific Control Representations}

% All control modalities are projected into a  $d$-dimensional conditioning space ($d=256$). 
% For each agent $i$, the modality-specific encoder produces a control embedding $z_i \in \mathbb{R}^d$. 
% Only targeted agents receive their corresponding $z_i$, while non-targeted agents are assigned $z_i = \mathbf{0}$.
% \paragraph{Context-Aware Behavior Latent Control.}

\subsubsection{Sketch and Language Conditioning.}
Trajectory sketches provide coarse spatial guidance in the form of future waypoints expressed in the agent’s local frame. 
We encode the waypoint sequence using a lightweight temporal encoder and aggregate it into a fixed-dimensional embedding $z_i$, capturing the intended motion trend while allowing reactive interactions.

Following~\cite{langtraj}, we utilize a DistilBERT~\cite{sanh2019distilbert} model adapted via Low-Rank Adaptation (LoRA)~\cite{lora} for natural language commands (e.g., ``turn left at the intersection'').
% For natural language commands (e.g., "turn left at the intersection"), we utilize a DistilBERT~\cite{sanh2019distilbert} model adapted via Low-Rank Adaptation (LoRA)~\cite{lora}. 
%
% \ychen{Why DistilBERT via LoRA? why not other approaches?} 
%
While language instructions are inherently abstract, we establish spatial and kinematic grounding by jointly optimizing the LoRA parameters with the downstream closed-loop control objective. Specifically, given a text prompt $p_i$, we extract the language features $\mathbf{H}_i = \mathrm{BERT}(p_i)$. We pool the \texttt{[CLS]} token embedding and apply a learned linear projection to match the shared $d$-dimensional control space.

\subsubsection{Context-Aware Behavior Latent.}

Beyond explicit sketches or language commands, we introduce a context-aware behavior latent that captures high-level driving patterns directly from data, without manual annotations. 
This compact control space abstracts interaction styles such as yielding or merging, enabling scalable and data-efficient controllability (see~\cref{subsec:sample_efficiency}).

\begin{figure}[t]
\centering
\includegraphics[width=0.95\linewidth]{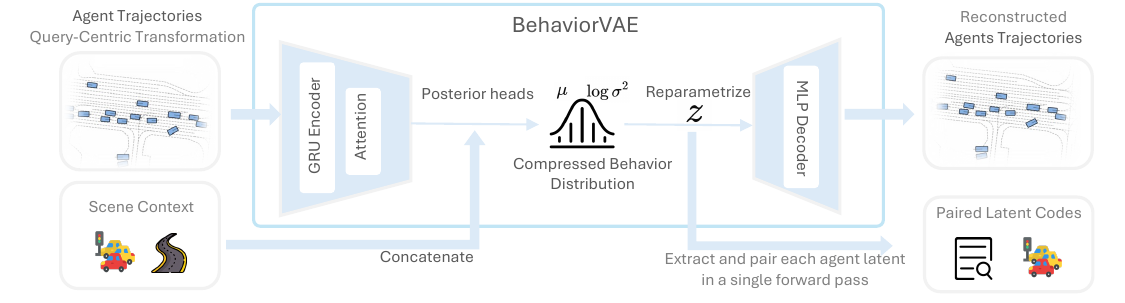}
\caption{\textbf{BehaviorVAE overview.} Agent trajectories and scene context are encoded into a Gaussian latent posterior; reparameterized per-agent latents are decoded for trajectory reconstruction and
exported as paired latent codes in a single forward pass.
% \ychen{I did not see this part in the introduction.} \yh{include in intro}
}
\label{fig:behaviorvae}
\end{figure}

To learn this representation, we employ a Conditional VAE (CVAE)~\cite{cvae}, referred to as \emph{BehaviorVAE}, which is trained separately from the control adapters.
We adopt a CVAE instead of deterministic autoencoder-\allowbreak based approaches~\cite{scenario_dreamer, realgen} because KL regularization encourages a smooth latent space where similar behaviors cluster together. This representation enables strong sample efficiency, as shown in \cref{subsec:sample_efficiency}.
Given a traffic scene, the encoder jointly processes all agents’ future motion sequences together with scene context in a single forward pass. 
By modeling relative geometry and multi-agent dependencies, it infers a latent variable $z_i$ for each agent that captures structured interaction behaviors in a context-aware manner (see~\cref{fig:behaviorvae}).

% \ychen{do you have a particular reason for choosing CVAE, instead of other approaches? Reviewers might ask. It is better to give justifications in the intro or here.} \yh{evidence in supplementory}
%

Formally, the encoder produces a Gaussian latent distribution 
$q_\phi(z_i \mid \boldsymbol{\xi}, c_{\mathrm{env}})$, 
where $\boldsymbol{\xi}$ denotes the set of agent trajectories in the scene. 
Because latent inference is performed jointly at the scene level, the resulting $z_i$ encodes interaction-aware behavior patterns rather than independent per-agent dynamics.
The model is trained using a conditional ELBO objective:
\begin{equation}
\mathcal{L}_{\mathrm{VAE}} 
= \mathbb{E}_{q}[\log p_\psi(\boldsymbol{\xi}_i \mid z_i)] 
- \beta D_{\mathrm{KL}}(q_\phi(z_i \mid \boldsymbol{\xi}, c_{\mathrm{env}}) \parallel p(z)),
\end{equation}
with KL annealing for stable optimization.

At inference time, sampling $z_i$ from the prior enables counterfactual behavior generation, and the latent is projected into the shared conditioning space for downstream control.

\subsection{Backbone Integration and Closed-Loop Fine-Tuning}
To enable controllable traffic simulation, the proposed adapters must be integrated into existing generative traffic models while preserving their learned generative priors. 
A key challenge is that the model operates in a closed-loop setting, where small prediction errors can accumulate over time and destabilize multi-agent interactions. 
We integrate control adapters into the generative cores of both diffusion~\cite{vbd} and autoregressive traffic models~\cite{smart} while keeping backbone parameters frozen.
%
% In both cases, control is injected at the decoder level, allowing the conditioning signal to directly influence the dynamics of trajectory generation. 
In both cases, control is injected at the decoder level to directly influence trajectory generation, systematically modulating the representations at every transformer decoder layer output across both the diffusion and autoregressive backbones.
%
%
% \ychen{Readers might wonder if you do not inject at every transformer decoder layer, what would be the corresponding difference? In addition, they might want to know what is the rationale of doing this? Why the proposed approach is better than other injection?} \yh{evidence in supplementory}
%
Only the control encoders and FiLM layers are updated during closed-loop fine-tuning \cite{catk,langtraj}. By leveraging FiLM-based feature modulation at the decoder level, the same control adaptation mechanism can be applied to different generative backbones without modifying their core architectures.

% \vspace{-4mm}

% \ychen{For the following two paragraphs, I think it would be better to remind readers that you want to explain how to finetune Diffusion and AR in a closed-loop manner. I could not fully follow the flow.}

% \yh{While the adaptation architecture is shared, we tailor the closed-loop fine-tuning strategies to accommodate the distinct generative paradigms of each backbone.}

The key idea of our closed-loop fine-tuning is to bias sampled trajectories toward those aligned with the control target while maintaining stable multi-agent interactions. 
The exact procedure depends on the generative backbone.
\paragraph{Diffusion: Receding-Horizon Control.}
Following a ``plan--select--execute'' paradigm~\cite{langtraj}, we generate \(M=8\) candidate rollouts at each replanning step and select the candidate closest to the ground-truth trajectory:
\begin{equation}
m^* = \operatorname*{argmin}_{m} 
\mathcal{L}_{\mathrm{match}}(\hat{\mathbf{S}}^{(m)}, \mathbf{S}^{gt}),
\end{equation}
where $\hat{\mathbf{S}}^{(m)}$ denotes the $m$-th predicted joint trajectory and $\mathbf{S}^{gt}$ is the ground truth. 
The adapter parameters are updated using the selected rollout to compute the training loss during closed-loop fine-tuning.

\paragraph{Autoregressive: Closed-Loop Fine-Tuning.}

For the autoregressive (AR) model, trajectories are generated sequentially under control conditioning $z_i$. 
We optimize a cross-entropy objective under closed-loop unrolling:
\begin{equation}
\mathcal{L}_{\mathrm{AR}} 
= - \frac{1}{NT} 
\sum_{i,t} 
\log \pi_\theta(s_{i,t}^{gt} \mid \hat{\mathbf{S}}_{\le t}, z_i),
\end{equation}
where $\hat{\mathbf{S}}_{\le t}$ denotes the dynamically generated joint state history up to time $t$, where $N$ is the number of agents and $T$ is the prediction horizon. To reinforce controllability, we bias sampling toward trajectories~\cite{catk} aligned with the conditioning signal, enabling the adapter to recover from its own prediction errors.
% To bridge the domain gap between training and inference, we employ scheduled sampling, actively mixing GT states with model predictions during the closed-loop rollout.

\subsection{Context-Aware Scenario Transfer}\label{sec:context_transfer}
In traffic simulation, it is important to explore alternative futures or synthesize rare interactions that deviate from recorded data while remaining realistic. However, directly forcing arbitrary control signals often leads to implausible behaviors in the current traffic context. To address this, we introduce a \textbf{context-aware scenario transfer pipeline} that retrieves compatible agents from similar scenarios and transfers their control signals into the target traffic scene, enabling realistic variations within existing environments.

\begin{figure*}[t!]
  \centering
  \includegraphics[width=0.95\linewidth]{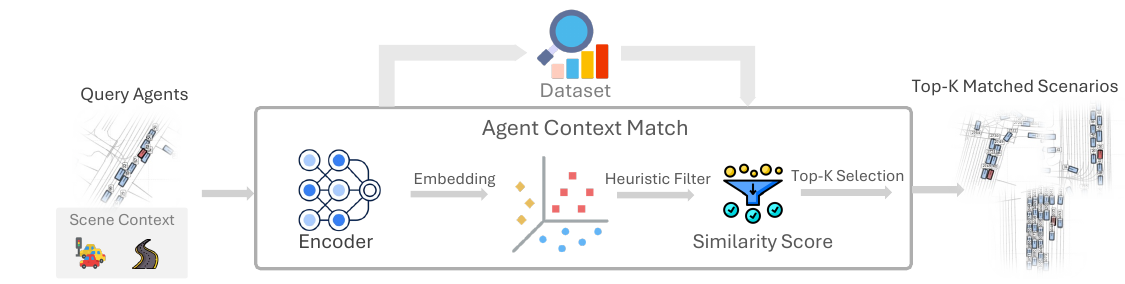}
  \caption{\textbf{Context-Match Retrieval Pipeline.} Query agents are encoded into a shared embedding space. Following heuristic filtering for dynamic feasibility, candidates are ranked via similarity scoring to retrieve the Top-K environmentally compatible scenarios from the dataset.}
  \label{fig:contextmatch}
\end{figure*}

Specifically, we use \emph{context embedding similarity} to retrieve compatible agents across scenarios. Given a target agent, we search the dataset for agents whose encoded scene context embeddings are similar (see~\cref{fig:contextmatch}). The retrieved agents serve as sources of plausible behaviors that can be transferred to the target scene.

Using this scenario transfer mechanism, we support two types of controllable scenario generation: 
\textbf{(i) Counterfactual Rollouts}, where we retrieve context-matched agents representing different driving intentions and transfer their behavior signals (e.g., sketch or latent trajectory) to the target scene to generate alternative futures, and 
\textbf{(ii) Long-Tail Synthesis}, where we identify rare behaviors using SMART likelihood and transfer their latent behavior codes to context-matched agents in new scenes to synthesize rare interactions.

% We propose a rigorous \emph{context-match retrieval} mechanism. To ensure environment plasubility, we first pre-filter candidates by taking the union of trajectory-level and scene context embedding nearest neighbors, strictly matching agent types. Candidates are then ranked using trajectory similarity, map topology constraints, and neighbor dynamics, bounded by strict feasibility gates. Leveraging this robust retrieval engine, we propose two complementary synthesis tasks:

%% file: sec/4_experiments.tex
\section{Experiments}
\label{sec:experiments}

% \ychen{It would be appreciated to have a summary of your experiments -- what do you want to prove and demonstrate?}
In this section, we demonstrate that our framework enables effective multi-modal controllability for both autoregressive and diffusion traffic models using only small amounts of paired control data, and further show its practical utility for counterfactual traffic simulation.

\subsection{Experimental Setup}
\label{subsec:exp_setup}

\noindent\textbf{Datasets and Benchmark.}
We evaluate our framework on the Waymo Open Motion Dataset (WOMD)~\cite{waymo}, following the standard WOSAC~\cite{wosac} protocol for closed-loop traffic simulation. BehaviorVAE is trained separately on 10,000 WOMD training scenarios.

\noindent\textbf{Backbone Models.}
To demonstrate the architecture-agnostic nature of our framework, we evaluate two state-of-the-art generative traffic models as unconditional backbones: \textbf{VBD}~\cite{vbd}, a diffusion-based model, and \textbf{SMART}~\cite{smart}, an autoregressive model.

\noindent\textbf{Metrics.}
Following established practices in controllable generation~\cite{prosim, langtraj}, we evaluate along two axes: 
\textbf{Realism}, measured by the WOSAC Meta score ($\uparrow$), evaluates distributional realism using likelihood-based metrics across multiple aspects of driving behavior. \textbf{Controllability}, measured by the minimum Average Displacement Error (mADE $\downarrow$) and relative mADE Gain ($\uparrow$), which quantify alignment with the intended control signals.
For counterfactual scenario synthesis (Sec.~\ref{subsec:latent_transfer}), we additionally report driving quality and safety metrics, including collision and off-road rates, as well as an adapted closed-loop \textbf{PDMScore}~\cite{navsim}.

\noindent\textbf{Control Signal Generation.}
To quantitatively evaluate controllability, we derive control signals from held-out ground truth (GT) trajectories: (i) \textbf{Sketch:} We downsample the GT future trajectories to obtain sparse waypoints as the sketch input; (ii) \textbf{Latent:} We process the GT trajectories through our pre-trained BehaviorVAE encoder to extract latent codes $\mathbf{z}$; and (iii) \textbf{Text:} We utilize natural language descriptions of each agent's motion from the \textbf{ProSim-Instruct-520k} dataset~\cite{prosim}. We explicitly note that derived controls are used only for standardized quantitative evaluation (e.g., mADE), not as a deployment assumption. Practical non-oracle controllability is evaluated via retrieval-based sketch/latent/text transfer in Sec.~\ref{subsec:latent_transfer}.

\subsection{Controllable Simulation Evaluation}
\label{subsec:main_results}
\cref{tab:main_results} reports controllable simulation results across both diffusion-based (VBD) and autoregressive (SMART) traffic models. To establish robust backbones for controllable simulation, we first convert the open-loop VBD and SMART models into closed-loop traffic models via additional training, denoted as \textbf{VBD-CL} and \textbf{SMART-tiny-CLSFT}~\cite{catk}. All control adapters in this comparison are trained using only \textbf{1\% of the paired control data}.
Despite this limited supervision, our adapters consistently reduce mADE and improve the overall WOSAC Meta score relative to the corresponding unconditional backbones across all three control modalities. Spatially grounded signals provide the strongest gains: \textbf{Sketch} and \textbf{Latent} reduce mADE by approximately \textbf{65\%} on VBD and by up to \textbf{83\%} on SMART. Text conditioning also improves controllability and overall realism, although its control gain is smaller because natural-language instructions provide weaker spatial grounding than trajectory sketches or behavior latents.
To isolate the effect of the proposed adaptation mechanism from backbone strength, we apply our FiLM adapters to a retrained unconditional ProSim backbone. As shown in \cref{tab:prosim_film_comparison}, using only \textbf{1\%} of the paired control data achieves sketch controllability comparable to the original ProSim model trained with \textbf{100\%} paired control data. For text conditioning, FiLM adaptation provides substantially stronger controllability despite using 100$\times$ less paired supervision. These same-backbone results show that the observed data efficiency comes from the proposed adaptation mechanism rather than solely from differences in backbone strength.\footnote{Original ProSim uses the official checkpoint; FiLM uses a retrained backbone (details in Supp. E3); gains are computed relative to each model's unconditional baseline.}

\begin{table}[t]
\centering
\caption{\textbf{Controllable Simulation Across Backbones and Modalities.}
Using only \textbf{1\%} of the paired control data, our lightweight FiLM adapters consistently reduce mADE and improve the overall WOSAC Meta score across both diffusion and autoregressive backbones. Bold indicates the best metric within each block.}
\label{tab:main_results}
\resizebox{0.95\linewidth}{!}{%
\setlength{\tabcolsep}{4pt}
\begin{tabular}{l c c c c c c c}
\toprule
\textbf{Backbone (Data Ratio)}
& \textbf{Trainable Params}
& \textbf{Meta} $\uparrow$
& \textbf{Kinematic} $\uparrow$
& \textbf{Interactive} $\uparrow$
& \textbf{Map} $\uparrow$
& \textbf{mADE} $\downarrow$
& \textbf{mADE Gain} $\uparrow$ \\ % <-- 修正
\midrule

\multicolumn{8}{l}{\cellcolor{gray!15}\textbf{Unconditional Base Models}} \\ % <-- 修正
VBD-CL
& 12.3M & 0.7186 & 0.4181 & 0.7705 & 0.8237 & 2.7223 & -- \\ % <-- 修正
SMART-tiny-CLSFT
& 7.0M & \textbf{0.7728} & \textbf{0.4660}
& \textbf{0.8057} & \textbf{0.9058} & \textbf{1.5044} & -- \\ % <-- 修正

\midrule
\multicolumn{8}{l}{\cellcolor{gray!15}\textbf{Sketch Control}} \\ % <-- 修正
VBD-CL (1\%)
& 1.3M & 0.7432 & 0.4720 & 0.7592 & 0.8776 & 0.9645 & 64.56\% \\ % <-- 修正
SMART-tiny-CLSFT (1\%)
& 2.1M & \textbf{0.7984} & \textbf{0.5305}
& \textbf{0.8146} & \textbf{0.9307}
& \textbf{0.2561} & \textbf{82.97\%} \\ % <-- 修正

\midrule
\multicolumn{8}{l}{\cellcolor{gray!15}\textbf{Latent Control}} \\ % <-- 修正
VBD-CL (1\%)
& 1.9M & 0.7440 & 0.4625 & 0.7801 & 0.8584 & 0.9221 & 66.12\% \\ % <-- 修正
SMART-tiny-CLSFT (1\%)
& 1.4M & \textbf{0.7912} & \textbf{0.5227}
& \textbf{0.8077} & \textbf{0.9234}
& \textbf{0.3238} & \textbf{78.48\%} \\ % <-- 修正

\midrule
\multicolumn{8}{l}{\cellcolor{gray!15}\textbf{Text Control}} \\ % <-- 修正
VBD-CL (1\%)
& 1.9M & 0.7306 & 0.4270 & 0.7670 & 0.8572
& 2.0005 & \textbf{26.51\%} \\ % <-- 修正
SMART-tiny-CLSFT (1\%)
& 2.0M & \textbf{0.7767} & \textbf{0.4751}
& \textbf{0.8072} & \textbf{0.9098}
& \textbf{1.3537} & 10.02\% \\ % <-- 修正

\bottomrule
\end{tabular}%
}
\end{table}

\begin{table}[t]
\centering
\caption{\textbf{Same-Backbone Comparison on ProSim.}
Using only \textbf{1\%} of the paired control data, FiLM adaptation achieves sketch controllability comparable to the original ProSim trained with \textbf{100\%} paired data, while providing stronger text controllability. The mADE gain is computed relative to each block's unconditional baseline.}
\label{tab:prosim_film_comparison}
\resizebox{0.7\linewidth}{!}{%
\begin{tabular}{l c c c c}
\toprule
\textbf{Method}
& \textbf{Paired Control Data}
& \textbf{Meta} $\uparrow$
& \textbf{mADE} $\downarrow$
& \textbf{mADE Gain} $\uparrow$ \\ % <-- 修正
\midrule

\multicolumn{5}{l}{\cellcolor{gray!15}\textbf{Original ProSim}} \\ % <-- 修正
ProSim Uncond.
& -- & 0.704 & 2.679 & -- \\ % <-- 修正
ProSim Sketch
& 100\% & 0.749 & \textbf{1.099} & \textbf{58.96\%} \\ % <-- 修正
ProSim Text
& 100\% & 0.709 & 2.330 & 13.03\% \\ % <-- 修正

\midrule
\multicolumn{5}{l}{\cellcolor{gray!15}\textbf{ProSim with FiLM Adaptation}} \\ % <-- 修正
ProSim Uncond.
& -- & 0.703 & 2.641 & -- \\ % <-- 修正
ProSim + FiLM Sketch
& 1\% & \textbf{0.750} & 1.130 & 57.23\% \\ % <-- 修正
ProSim + FiLM Text
& 1\% & \textbf{0.717} & \textbf{1.824} & \textbf{30.92\%} \\ % <-- 修正

\bottomrule
\end{tabular}%
}
\end{table}

% Despite training our control adapters with only \textbf{1\%} of the dataset and keeping the backbone models frozen, our approach achieves strong controllability improvements while preserving simulation realism.

% Across both generative families, the \textbf{Sketch} and \textbf{Latent} modalities achieve profound improvements in controllability, demonstrating strict alignment with user-defined spatial and behavioral intents. Crucially, this precise steering enhances, rather than disrupts, the overall simulation realism (Meta scores), suggesting that valid future guidance resolves generative ambiguity. Furthermore, our \textbf{Text} adapter successfully imparts high-level semantic control. While natural language inherently provides sparser spatial grounding compared to explicit trajectories or behavior codes, it still yields consistent controllability gains and preserves stable closed-loop interactions, even when steering the notoriously rigid AR backbone. To contextualize our performance, we compare against ProSim~\cite{prosim}, the most closely related baseline providing sketch and text control. As shown in Table~\ref{tab:main_results}, ProSim relies on full-parameter updates using 100\% of the dataset. Remarkably, our lightweight adapters, utilizing strictly frozen backbones and only 1\% of the data, achieve superior relative performance. 

\begin{figure}[t]
  \centering
  \includegraphics[width=0.9\linewidth]{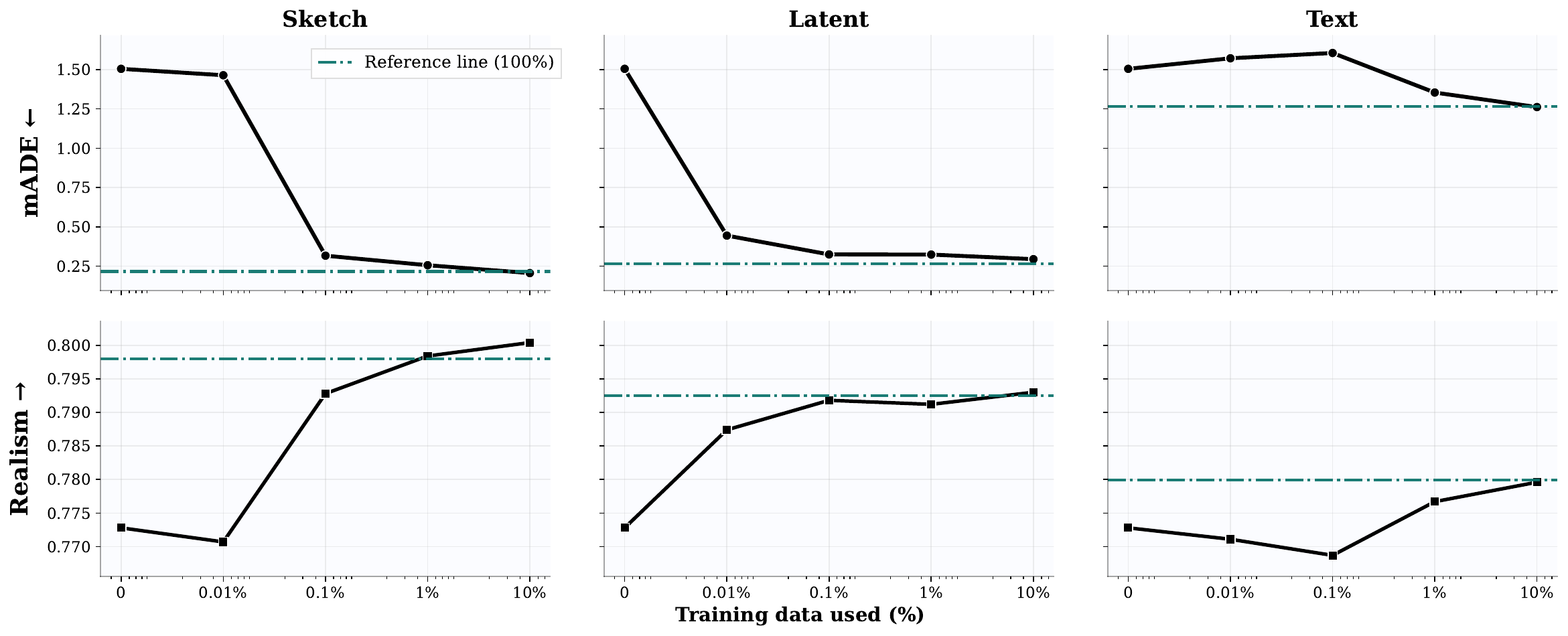}
  \caption{\textbf{Sample Efficiency on WOSAC.} Evaluated using the autoregressive backbone. We report controllability (mADE $\downarrow$, top) and realism (Meta Score $\uparrow$, bottom) as a function of training data size. The unconditional base model corresponds to the $0\%$ data point. Our adapters achieve strong controllability with minimal supervision, surpassing the base model with only \textbf{0.01\%} (Latent) and \textbf{0.1\%} (Sketch) data while maintaining comparable realism. Performance saturates around 10\% data, matching a full-data LoRA finetuning.}
  \label{fig:sample_efficiency}
\end{figure}
%take away ehre, latent is a good modaltity even though 0.1 percent data can learn a good 
\subsection{Sample Efficiency and Data Scaling}
\label{subsec:sample_efficiency}
In this section, we focus on evaluating the \textbf{sample efficiency} of our proposed adaptation framework. Adapters are trained on logarithmically spaced subsets of the Waymo paired control dataset (50 to 50,000 scenarios; 0.01\% $\sim$ 10\% data).
Figure~\ref{fig:sample_efficiency} illustrates how controllability (mADE $\downarrow$) and distributional realism (Meta Score $\uparrow$) evolve as the amount of training data increases. 
For reference, we additionally report the performance obtained by fine-tuning the base model with LoRA on the full $100\%$ dataset using the same control adapters.

\paragraph{Data Efficiency Across Modalities.}
Our adapters demonstrate strong sample efficiency across control modalities. 
Even with extremely small amounts of paired control data, they are able to learn effective controllable behaviors while preserving the generative prior of the frozen base model. Among the evaluated modalities, \textbf{Latent} control is the most data-efficient: with only \textbf{0.01\%} of the training data, it already achieves performance close to the reference model trained with the full dataset. The \textbf{Sketch} modality also performs strongly under limited supervision, approaching the reference performance with merely \textbf{0.1\%} of the data. In contrast, the \textbf{Text} modality exhibits a delayed scaling curve, requiring ${\sim}1\%$ data to outperform the baseline. Because natural language provides abstract guidance and relies on noisy auto-generated captions (e.g., ProSim-Instruct-520k), text control primarily serves as high-level intent biasing rather than precise trajectory supervision.

\paragraph{Realism Preservation and Saturation.}
Importantly, extremely low-data regimes do not compromise simulation realism. 
While Meta scores for Sketch and Text dip slightly at $0.01\%$, they rapidly recover and match the base model by $0.1\%$ data, while Latent control improves realism from the start. Performance across all modalities saturates around $10\%$ data, where Sketch, Latent and Text adapters achieve mADE comparable to the \textbf{reference performance} obtained by LoRA fine-tuning on the full $100\%$ dataset (see~\cref{fig:sample_efficiency}). 
This suggests that effective controllable simulation can be achieved without large-scale paired annotations.

\subsection{Counterfactual and Long-Tail Scenario Generation}
\label{subsec:latent_transfer}

Having established the sample efficiency of our control adaptation framework, we next evaluate its practical utility for scenario generation using the context-aware scenario transfer pipeline described in \cref{sec:context_transfer}. Specifically, we generate counterfactual and long-tail scenarios within existing traffic scenes. For these experiments, we use logits-level classifier-free guidance during decoding; see Section E.1 of the supplementary material.

\begin{table}[t]
\centering
\caption{\textbf{Counterfactual Rollouts.} 
Control adapters enable diverse counterfactual rollouts while maintaining closed-loop driving quality. 
Sketch and latent controls achieve strong controllability (low Control ADE and high maneuver success), while all modalities increase trajectory diversity (Coverage) compared to the unconditional model without degrading safety metrics or PDMScore.}
\label{tab:counterfactual_transfer}
\resizebox{0.95\linewidth}{!}{%
\setlength{\tabcolsep}{4pt}
\begin{tabular}{l c c c c c c c}
\toprule
\textbf{Method} & \textbf{Control ADE (m)} & \textbf{Success (\%)} $\uparrow$  & \textbf{Coverage} $\uparrow$ & \textbf{Coll.} $\downarrow$ & \textbf{Offroad} $\downarrow$ & \textbf{PDMScore} $\uparrow$ \\
\midrule
Unconditional  & -- & --  & 201.02 & 0.0563 & 0.0252 & 78.40 \\
\textbf{Sketch CF} & 0.3727 & 70.61 & \textbf{287.17} & 0.0836 & 0.0252 & 79.30 \\
\textbf{Latent CF} & 0.4542 & \textbf{73.89} & 254.41 & 0.0774 & 0.0251 & \textbf{80.33} \\
\textbf{Text CF} & 2.4446 & 69.84  & 282.57 & \textbf{0.0520} & \textbf{0.0227} & 79.56 \\
\bottomrule
\end{tabular}%
}
\end{table}

\noindent(i) \textbf{Counterfactual Rollouts:} We select 250 query agents and retrieve context-matched agents representing common driving intentions (\emph{turning, lane changing, accelerating, decelerating, cruising, stopping}). 
For each query agent, we select up to four distinct intentions with the highest context-match scores and transfer the corresponding behavior annotations (\emph{sketch} or \emph{latent}) to the query agent in the original scene to generate counterfactual rollouts.

\textbf{Evaluation Metrics.}
For counterfactual generation, we evaluate three aspects: 
(i) \textbf{controllability}, i.e., how well the transferred behavior controls the target agent, 
(ii) \textbf{behavioral diversity}, and 
(iii) \textbf{driving realism}.  
Controllability is measured using \textbf{Control ADE}, which computes the displacement error between the generated trajectory and the transferred control behavior (e.g., sketch or latent trajectory). 
We also report the maneuver \textbf{Success Rate}, defined as the fraction of rollouts that successfully execute the same high-level maneuver in each scenario.
To evaluate diversity, \textbf{Coverage} measures the spatial distributional spread of generated trajectories, computed as the number of grid cells whose kernel density estimate (KDE) exceeds a predefined threshold~\cite{bits}.
Driving realism is evaluated using \textbf{Collision}, \textbf{Offroad}, and the closed-loop \textbf{PDMScore}~\cite{navsim}. We do not report WOSAC metrics in this setting, as they measure similarity to ground-truth trajectories, which is not appropriate for counterfactual generation where trajectories intentionally deviate from the recorded future.

Table~\ref{tab:counterfactual_transfer} shows that our control adapters enable effective counterfactual generation across modalities. 
Sketch and latent controls achieve strong controllability, reflected by low Control ADE and high maneuver success rates (~70\%). 
At the same time, driving realism is preserved, with all modalities maintaining stable safety metrics and achieving PDM scores comparable to or higher than the unconditional baseline. 
Finally, the generated trajectories exhibit greater diversity: conditioning on different high-level behaviors increases trajectory coverage compared to the unconditional baseline, indicating that our framework can explore multiple plausible futures within the same scene (see Sec.~A of the supplementary material for qualitative results).

\begin{figure*}[t!]
  \centering
  \includegraphics[width=0.8\linewidth]{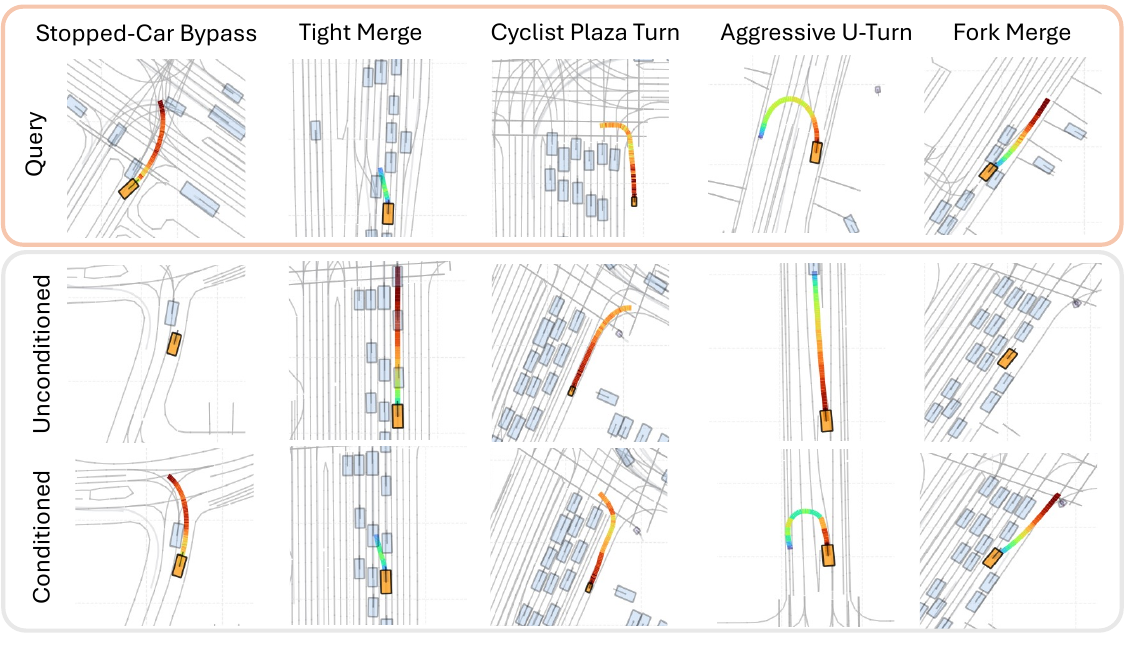}
  \caption{\textbf{Qualitative results of long-tail scenario generation.} 
  \textbf{Top:} Source scenarios providing the target long-tail behavioral latents. 
  \textbf{Middle:} Default base model predictions in the retrieved matching contexts, showing only nominal driving. 
  \textbf{Bottom:} Our framework successfully injects the queried intent, enabling agents to execute complex, safety-critical maneuvers realistically in novel contexts.
  % \ychen{how fast is your inference and retrieval? can you make this into a UI so that users can random queries?} \textcolor{blue}{retrieval is offline processed, inference is maybe 1min?, yes UI is possible with behvior speciifed query}
  }
  \label{fig:longtail}
\end{figure*}

\noindent(ii) \textbf{Long-Tail Synthesis:} 
Beyond generating alternative futures within a scene, we also evaluate whether our framework can synthesize rare behaviors in new contexts. We use the SMART model's likelihood from~\cite{catk} to identify the top $0.1\%$ hardest-to-predict scenarios in the dataset. 
From these scenarios, we curate eight long-tail behavior categories, including \emph{plaza turning, meandering, threading stopped cars, multi-lane weaving, unsignalized crossing, fork-merge turning, tight merging,} and \emph{aggressive U-turns}, as shown in \cref{fig:longtail}.

Similar to the counterfactual setup, for each behavior, we retrieve the top 30 context-matched candidate agents and transfer the corresponding behavior latent $\mathbf{z}_{\text{target}}$ to the selected agent while leaving surrounding agents unconditioned. 
As a baseline, we compare against randomly selecting agents within the scene to receive the transferred behavior. 
Since long-tail behaviors intentionally deviate from nominal driving, we additionally report \textbf{Traj ADE} to quantify the extent to which the generated trajectory diverges from the unconditional simulation.

Table~\ref{tab:longtail_transfer} highlights the importance of context-aware transfer. 
Random context transfer often places the target behavior in incompatible environments, leading to significantly higher collision rates ($0.3175$) and degraded driving performance. 
In contrast, \textbf{Context Match} better aligns the injected behavior with compatible traffic contexts, reducing collisions ($0.1574$) and improving the PDMScore.

\begin{table}[t!]
\centering
\caption{\textbf{Long-Tail Scenario Generation.} 
Context-aware retrieval significantly improves intent alignment and driving quality when transferring rare behaviors to new contexts. 
Compared to random context selection, context matching reduces collisions and improves closed-loop driving performance.}
\label{tab:longtail_transfer}
\resizebox{0.8\linewidth}{!}{%
\setlength{\tabcolsep}{3pt}
\renewcommand{\arraystretch}{0.8}
\begin{tabular}{l c c c c c}
\toprule
\textbf{Strategy} & \textbf{Control ADE} $\downarrow$ & \textbf{Traj ADE} & \textbf{Coll.} $\downarrow$ & \textbf{Offroad} $\downarrow$ & \textbf{PDMScore} $\uparrow$ \\
\midrule
Context Random & 3.8723 & 16.33 & 0.3175 & 0.1553 & 39.06 \\
\textbf{Context Match} & \textbf{1.6994} & {11.30} & \textbf{0.1574} & \textbf{0.0661} & \textbf{65.27} \\
\bottomrule
\end{tabular}%
}
\end{table}

\begin{figure}[t]
  \centering
  \includegraphics[width=0.9\linewidth]{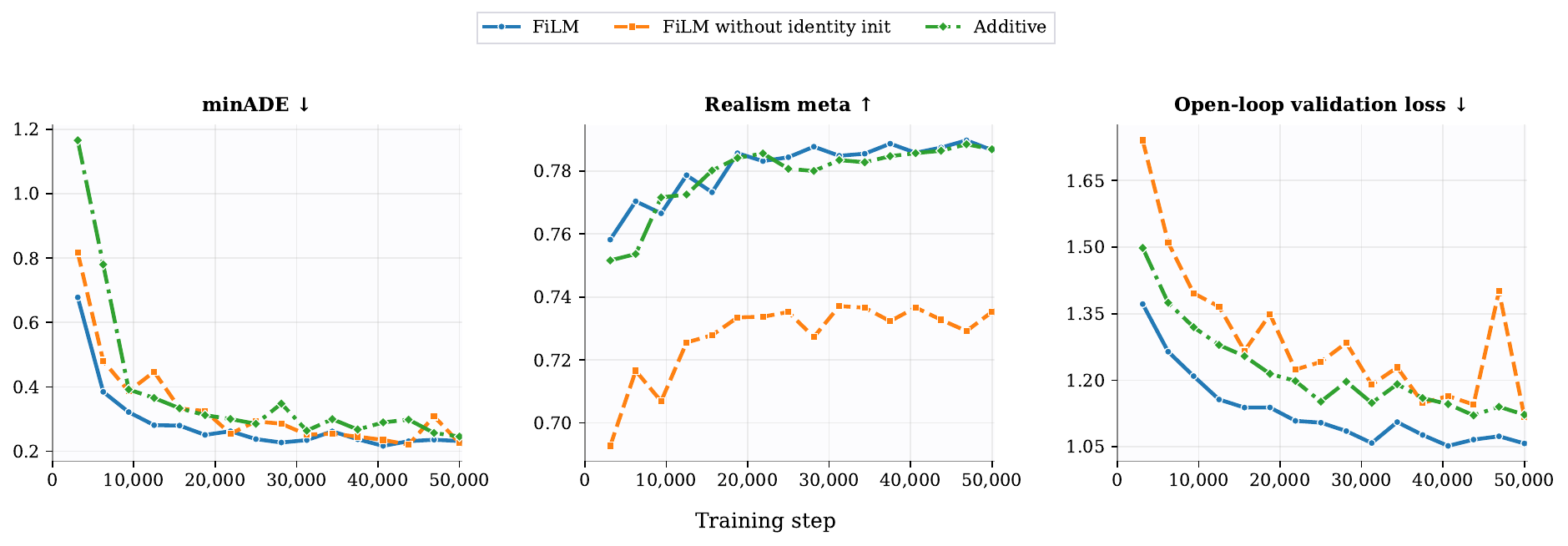}
  \caption{\textbf{Ablation on Modulation Mechanisms.} Multiplicative FiLM (Blue) converges faster and achieves lower mADE than additive modulation (Green), reaching strong performance within fewer optimization steps. Identity initialization is crucial for preserving the base model’s realism prior (Center) at the start of training.}
  \label{fig:ablation_modulation}
\end{figure}

Nevertheless, transferring long-tail behaviors across scenarios remains inherently challenging, as these maneuvers (e.g., tight merges) are rare and highly constrained. 
We observe that many collision cases result from the trade-off between enforcing the conditioned long-tail behavior and preserving safe interactions with surrounding agents. Finally, the large trajectory deviation (\textbf{Traj ADE} $=11.30$m) indicates that the model successfully overrides the nominal unconditional behavior to execute the targeted long-tail interactions.

\subsection{Ablation Studies}
\label{subsec:ablation}

We conduct additional ablations to validate our core design choices regarding adapter modulation design and conditioning strategies. Unless otherwise specified, models are trained on a 5,000-scenario subset ($\sim$1\% of WOMD) utilizing the autoregressive backbone to ensure rapid iteration while preserving statistical significance.

\subsubsection{Adapter Modulation Design.}
\label{subsec:control_modulation}

We study the architectural design of the control adapter using the Sketch modality. 
Figure~\ref{fig:ablation_modulation} compares our proposed \textbf{FiLM with Identity Initialization} against two baselines: FiLM with random initialization and additive modulation. 
The additive baseline injects control signals via bias addition, i.e., 
$\mathbf{h}^{(l)}_{\text{add}} = \mathbf{h}^{(l)} + \boldsymbol{\beta}^{(l)}(z_i)$.

As shown in~\cref{fig:ablation_modulation}, \textbf{identity initialization} is critical. 
Randomly initialized adapters significantly disrupt the pretrained model early in training, causing a severe drop in realism, whereas identity-initialized FiLM preserves the base model’s behavior from the start. 
Moreover, FiLM is substantially more efficient: while additive modulation fails to converge after 50k steps ($\sim$9 hours on a single RTX 4090), our FiLM adapter reaches better mADE in under 12k steps ($\sim$2.2 hours). 
These results suggest that identity-initialized multiplicative modulation provides a stronger inductive bias for control-adapter fine-tuning.

\begin{table}[t]
\centering
\caption{\textbf{Static vs. Dynamic Control Conditioning.}
Realism and controllability results for Sketch and Latent modalities.}
\label{tab:static_dynamic}
\resizebox{0.9\linewidth}{!}{%
\setlength{\tabcolsep}{5pt}
\renewcommand{\arraystretch}{0.8}
\begin{tabular}{l c c c c c c}
\toprule
\textbf{Representation} & \textbf{Time Emb.} & \textbf{Meta} $\uparrow$ & \textbf{Kinematic} $\uparrow$ & \textbf{Interactive} $\uparrow$ & \textbf{Map} $\uparrow$ & \textbf{mADE} $\downarrow$ \\
\midrule
\multicolumn{7}{l}{\textit{Sketch Control Modality}} \\
\quad \textbf{Dynamic} & $\times$ & 0.7799 & 0.5261 & 0.7921 & 0.9091 & \textbf{0.3970} \\
\quad Static & \checkmark & 0.7719 & 0.5083 & 0.7823 & 0.9091 & 1.0598 \\
\midrule
\multicolumn{7}{l}{\textit{Latent Control Modality}} \\
\quad \textbf{Dynamic } & $\times$ & 0.7848 & 0.5272 & 0.8027 & 0.9089 & \textbf{0.3151} \\
\quad Static & \checkmark & 0.7880 & 0.5113 & 0.8087 & 0.9195 & 0.4842 \\
\bottomrule
\end{tabular}
}
\end{table}

% \subsubsection{Control Conditioning Strategy.}
% \label{subsec:static_dynamic}

% We compare two conditioning strategies for Sketch and Latent signals: 
% (i) \textbf{Static}, where the future trajectory is encoded once and injected with a time embedding across steps, and 
% (ii) \textbf{Dynamic}, where the control signal is re-encoded at each timestep to follow the receding horizon.

% Table~\ref{tab:static_dynamic} shows that \textbf{latent control is effective}, achieving strong controllability under both strategies while maintaining high realism (e.g., Meta $=0.7880$, Map $=0.9195$ for static latent). 
% While dynamic conditioning achieves slightly lower mADE ($0.3151$ vs.\ $0.4842$), \textbf{static conditioning remains competitive} and avoids enforcing strict timestep alignment in closed-loop interactions. This flexibility is desirable in closed-loop simulation, where agents must adapt their behavior in response to interactions with surrounding agents.

\subsubsection{Control Conditioning Strategy.}
\label{subsec:static_dynamic}

We compare two conditioning strategies:
(i) \textbf{Static}, which encodes the control signal once and injects it with a time embedding, and
(ii) \textbf{Dynamic}, which re-encodes the control signal at each timestep.

As shown in Table~\ref{tab:static_dynamic}, dynamic conditioning achieves lower mADE, while static conditioning preserves competitive realism and avoids strict timestep alignment. This flexibility is beneficial in closed-loop simulation, where agents must adapt to surrounding interactions.

% Based on this observation, we adopt a hybrid strategy: dynamic conditioning for dense spatial signals (Sketch) to maximize precision, and static conditioning for Latent and Text signals that represent higher-level behavioral intent.

%% file: sec/5_disc.tex
\section{Limitations and Future Work}
\label{sec:limitations}

Despite these gains, our method still faces limitations in handling infeasible or ambiguous controls. When a user-specified instruction conflicts with the surrounding traffic context, the model may prioritize control satisfaction over realistic and safe interactions. Future work could address this through RL fine-tuning or guidance-based inference to better balance control fidelity and realism. In addition, our experiments focus primarily on controlling a single agent, and future work will extend the framework toward coordinated multi-agent control.

\section{Conclusion}
\label{sec:conclusion}

We present a {data-efficient control adaptation framework} that enables controllable traffic simulation across multiple conditioning modalities, including sketch, latent behavior codes, and natural language. 
Built on top of pretrained generative traffic models, our lightweight adapters can be applied to both \textbf{autoregressive} and \textbf{diffusion-based} architectures without modifying the backbone models. Experiments show that our approach achieves strong controllability while preserving simulation realism using only \textbf{1\% of the paired control data}. Beyond standard control evaluations, our framework enables \textbf{counterfactual scenario generation} and \textbf{long-tail behavior synthesis}, producing diverse yet realistic traffic interactions.

% Future work includes planner-in-the-loop evaluation and extending the framework to support vulnerable road user simulation.

% We present a unified, lightweight control adaptation framework that endows frozen traffic simulators, including inherently rigid autoregressive models, with precise multi-modal controllability. Using under 2.1M parameters via identity-initialized FiLM layers, our approach robustly synthesizes safety-critical long-tail events and diverse counterfactuals while preserving closed-loop realism. Crucially, we demonstrate that robust control does not necessitate massive paired annotations; anchoring a strong generative prior with minimal, high-quality data proves highly effective. Building on this sample-efficient foundation, future work may leverage high-fidelity but relatively small datasets such as HetroD~\cite{hetrod} to further advance this paradigm toward heterogeneous traffic simulation. This is particularly important for modeling Vulnerable Road Users (VRUs), enabling more coherent interactions among diverse agents and more faithfully representing the complexity of real-world autonomous driving.

%% file: sec/6_suppl.tex
% \documentclass{article}
% \usepackage[utf8]{inputenc}
% \usepackage{amsmath, amssymb, amsthm}
% \usepackage{graphicx}
% \usepackage{booktabs} 
% \usepackage{hyperref}
% \usepackage{geometry}
% \usepackage{cleveref}
% \geometry{margin=1in}
% \usepackage{orcidlink}
% \usepackage{multirow}
% \usepackage{xr}
% \usepackage{tcolorbox} 
% \externaldocument{main} % 讀取 main.tex 的編號記錄檔
% % --- 頁碼與格式設定 (請勿更動) ---
% \clearpage
% \pagenumbering{gobble}
% \setcounter{page}{1}
% \setcounter{section}{0}
% \setcounter{figure}{0}
% \setcounter{table}{0}
% \setcounter{equation}{0}
% \setcounter{footnote}{0}
% \renewcommand{\thepage}{A\arabic{page}}
% \renewcommand{\thesection}{\Alph{section}}
% \renewcommand{\thefigure}{A\arabic{figure}}
% \renewcommand{\thetable}{A\arabic{table}}
% \renewcommand{\theequation}{A\arabic{equation}}

% \title{Supplementary Material \\ 
%  ECoSim: Data-Efficient Fine-Tuning for Controllable Traffic Simulation}
% \author{ECCV 2026 Submission \#2701}
% \date{}

% \newcommand{\yh}[1]{\textcolor{orange}{[YH]: #1}}

% \begin{document}
% \maketitle

% =========================================================
% Supplementary Material
% This file is included by main.tex using:
% \input{sec/6_suppl}
% =========================================================

\setcounter{section}{0}
\setcounter{figure}{0}
\setcounter{table}{0}
\setcounter{equation}{0}
\setcounter{footnote}{0}

\renewcommand{\thesection}{\Alph{section}}
\renewcommand{\thesubsection}{\thesection.\arabic{subsection}}
\renewcommand{\thefigure}{A\arabic{figure}}
\renewcommand{\thetable}{A\arabic{table}}
\renewcommand{\theequation}{A\arabic{equation}}

\begin{center}
    {\LARGE\bfseries Supplementary Material}\\[0.5em]
    {\normalsize ECoSim: Data Efficient Fine-Tuning for Controllable Traffic Simulation}
\end{center}

Compared to prior work, our approach enables data-efficient finetuning to introduce multiple control knobs for generating counterfactual behaviors in closed-loop simulation. Section \ref{sec:app_qualitative} provides qualitative results, Section \ref{sec:app_ablations} presents additional baseline comparisons and ablation studies, Section \ref{sec:app_limitations} provides discussion and limitations, Section \ref{sec:app_setup_metrics} details the experimental setup and metrics, and Section \ref{sec:app_implementation} covers guidance strategies, architecture and implementation details.

\begin{figure}[ht]
    \centering
    \includegraphics[width=\linewidth]{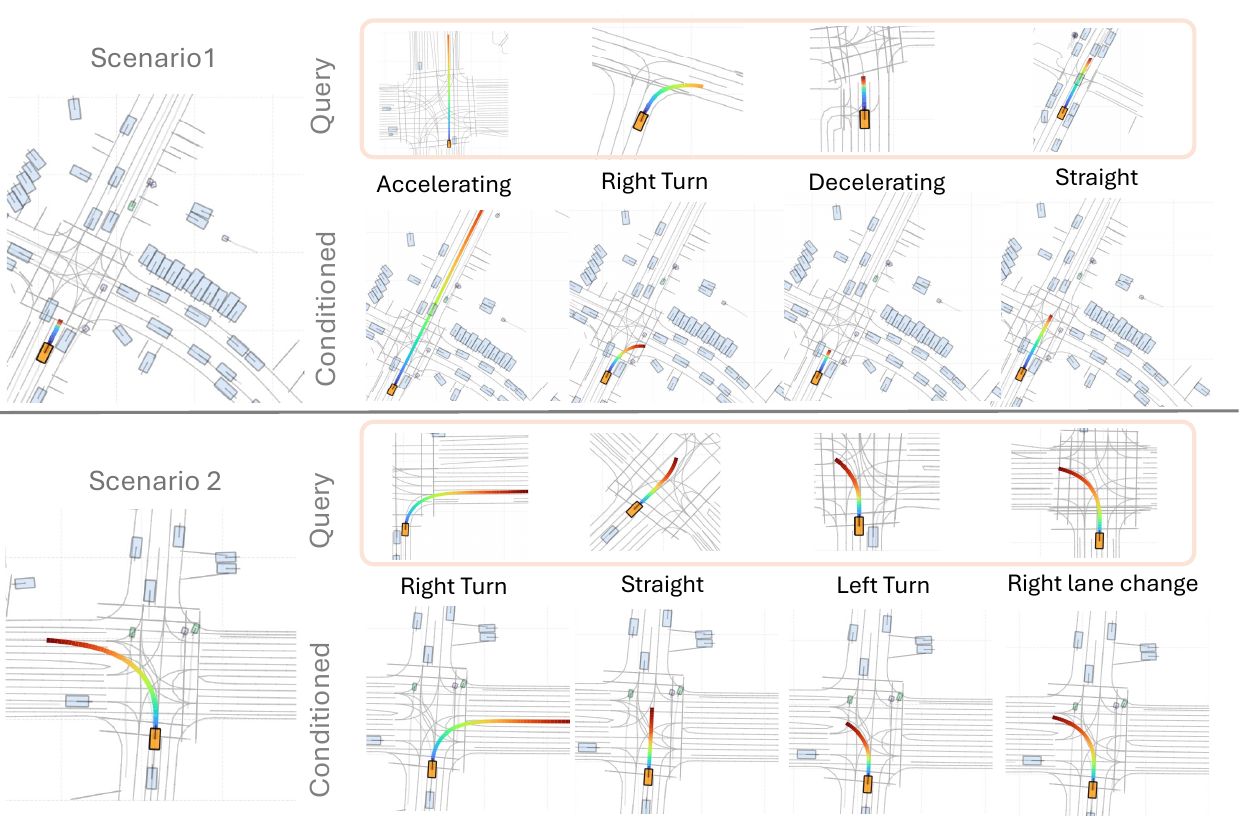}
    \caption{Qualitative counterfactual generation across different traffic scenarios. For each scenario, the \textbf{Query} row displays the isolated reference behaviors (e.g., accelerating, right turn, lane change), while the \textbf{Conditioned} row illustrates our model's synthesized trajectories. The results demonstrate that our framework successfully injects diverse control queries while seamlessly adapting the resulting kinematics to the target scene's unique road topology and interactive context.}
    \label{fig:counterfactual}
\end{figure}

% ==========================================
% D. 質性結果與討論
% ==========================================
\section{Qualitative Results}
To better illustrate the dynamic closed-loop performance of our framework, we invite reviewers to view the supplementary videos, which showcase complex counterfactual and long-tail simulation scenarios, visualizations shown in Fig.~\ref{fig:counterfactual}.

\label{sec:app_qualitative}

\section{Additional Experimental Results}
\label{sec:app_ablations}

\subsection{ProSim Baseline and Counterfactual Setup}
\label{subsec:prosim_counterfactual}

\paragraph{Evaluation Setup} 
We additionally report counterfactual generation results compared with ProSim~\cite{prosim}. Evaluation is conducted on the subset of validation scenarios that match ProSim’s processed data. Metrics are computed exclusively on the target controlled agents across all generated rollouts. The \textit{ProSim Unconditional (Uncond)} setting serves as a baseline representing the model’s unguided prior, where all condition inputs are fully masked.

\paragraph{Results and Comparison} 
As detailed in Table~\ref{tab:prosim_counterfactual}, ProSim suffers a performance trade-off when explicit controls (\textit{Sketch} or \textit{Text}) are introduced: while it attempts to follow the instructions, its overall simulation quality degrades (e.g., Text PDMScore drops to 70.25) and off-road rates increase. In contrast, \textbf{our framework (using the SMART~\cite{smart} backbone trained on a 10\% subset of WOMD) demonstrates transfer fidelity and generative diversity.} Our model achieves lower Control ADE across all modalities, translating to higher Success Rates ($\sim$87\% vs. 67-79\%). Crucially, our approach enforces control conditions without compromising base realism, maintaining high PDMScores ($\sim$77-78) and lower off-road rates ($\sim$0.040). Furthermore, our method generates substantially broader Spatial Coverage (up to 148.70 vs. ProSim's 105.60), proving it can execute specific prompts while exploring a richer diversity of plausible futures.

\begin{table}[ht]
    \centering
    \caption{Counterfactual evaluation on the matched ProSim subset. Compared to the ProSim baseline, our proposed framework achieves higher Success Rates and Spatial Coverage while maintaining superior trajectory fidelity (Control ADE) and overall simulation quality (PDMScore) across diverse control modalities.}
    \label{tab:prosim_counterfactual}
    \resizebox{\textwidth}{!}{
    \begin{tabular}{llcccccc}
        \toprule
        \textbf{Method} & \textbf{Setting} & \textbf{Control ADE (m)} $\downarrow$ & \textbf{Success (\%)} $\uparrow$ & \textbf{Coverage} $\uparrow$ & \textbf{Coll.} $\downarrow$ & \textbf{Offroad} $\downarrow$ & \textbf{PDMScore} $\uparrow$ \\
        \midrule
        \multirow{3}{*}{ProSim} 
        & Uncond      & -- & -- & 91.25 & \textbf{0.065} & 0.055 & \textbf{78.90} \\
        & Sketch Only & 0.728 & 79.58 & 92.41 & 0.107 & 0.081 & 72.35 \\
        & Text Only   & 1.768 & 67.14 & 105.60 & 0.133 & 0.077 & 70.25 \\
        \midrule
        \multirow{4}{*}{\textbf{Ours}} 
        & Uncond      & -- & -- & 91.25 & 0.072 & 0.040 & 76.46 \\
        & Sketch Only & \textbf{0.213} & 87.93 & \textbf{148.70} & 0.109 & 0.040 & 77.95 \\
        & Latent Only & 0.217 & \textbf{88.09} & 143.59 & 0.108 & \textbf{0.039} & 77.47 \\
        & Text Only   & 1.363 & 87.20 & 140.39 & 0.072 & 0.040 & 77.11 \\
        \bottomrule
    \end{tabular}
    }
\end{table}

\subsection{Target-Only vs. Full-GT Control Conditioning}
\label{subsec:target_agent_only}

We further examine the control-conditioning scope used in Sec.~4.2 of the main paper. Here, \emph{target agents} refer to the agents designated by the WOSAC protocol for evaluation in each WOMD scenario. Under the Full-GT setting, control signals are provided to all ground-truth-valid agents. Under the Target-Only setting, control signals are provided only to the WOSAC-designated target agents, while all other agents receive zero conditioning. Both settings are evaluated using the same metrics and evaluation protocol. As shown in Table~\ref{tab:target_only_ablation}, restricting control conditioning to the designated target agents leads to a reduction in control precision, particularly for Sketch control, while the realism metrics remain largely comparable. These results indicate that our method remains effective under target-only conditioning, although conditioning additional surrounding agents can further improve control accuracy and interaction consistency.

\begin{table}[t]
\centering
\caption{\textbf{Ablation on control conditioning scope.}
Metrics are reported under the same evaluation protocol, while control signals are provided either to all ground-truth-valid agents (Full GT) or only to the WOSAC-designated target agents (Target Only). Bold indicates the better value between Sketch and Text within each conditioning scope.}
\label{tab:target_only_ablation}
\resizebox{0.8\linewidth}{!}{%
\begin{tabular}{llccccc}
\toprule
Control Conditioning Scope & Control & Meta $\uparrow$ & Kinematic $\uparrow$
& Interactive $\uparrow$ & Map $\uparrow$ & mADE $\downarrow$ \\ % <-- 修正這裡
\midrule
\multirow{2}{*}{Full GT}
& Sketch & \textbf{0.7984} & \textbf{0.5305}
& \textbf{0.8146} & \textbf{0.9307} & \textbf{0.2561} \\ % <-- 修正這裡
& Text & 0.7767 & 0.4751 & 0.8072 & 0.9098 & 1.3537 \\ % <-- 修正這裡
\midrule
\multirow{2}{*}{Target Only}
& Sketch & \textbf{0.7898} & \textbf{0.5237}
& 0.8025 & \textbf{0.9254} & \textbf{0.4602} \\ % <-- 修正這裡
& Text & 0.7770 & 0.4767 & \textbf{0.8074} & 0.9095 & 1.4101 \\ % <-- 修正這裡
\bottomrule
\end{tabular}%
}
\end{table}

\subsection{Ablation on BehaviorVAE Posterior and Latent Space Analysis }
\label{subsec:latent_analysis}

To validate the BehaviorVAE's posterior formulation using the VBD~\cite{vbd} backbone, we investigate the impact of concatenating scene context embeddings with encoded trajectories before the CVAE posterior head ($\mu, \log \sigma^2$). 

\begin{figure}[ht]
    \centering
    \includegraphics[width=\linewidth]{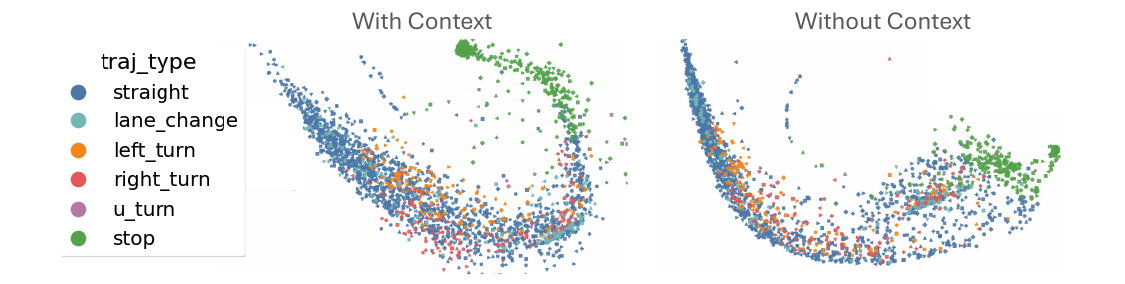}
    \caption{t-SNE visualization of the BehaviorVAE latent space. Colors indicate distinct semantic trajectory types. \textbf{Left:} The \textit{With Context} formulation, which concatenates scene context embeddings prior to the posterior head, captures rich behavioral diversity (intra-cluster variance) while maintaining clear semantic groupings. \textbf{Right:} The \textit{Without Context} posterior relies solely on trajectory features, producing overly concise clusters that lack the contextual variance necessary for topology-aware conditional generation.}
    \label{fig:latent_tsne}
\end{figure}

As shown in Table~\ref{tab:ablation_context}, the \textit{Without Context} baseline achieves a slightly higher Meta metric (0.7361 vs. 0.7335) by defaulting to conservative, generic kinematics that easily satisfy rigid rule-based safety metrics. However, explicitly conditioning on scene context (\textit{With Context}) improves spatial fidelity, reducing the Minimum Average Displacement Error (mADE) to 1.0228 by adapting generative predictions to specific road topologies. This quantitative trade-off is visually corroborated by our t-SNE~\cite{tsne} projections (Figure~\ref{fig:latent_tsne}). Relying strictly on raw kinematics (\textit{Without Context}) collapses latents into rigidly separated clusters. In contrast, concatenating scene context introduces crucial intra-cluster variance while maintaining clear semantic manifolds (e.g., straight, stop, turns). This context-induced diversity prevents CVAE mode collapse, providing the generative flexibility necessary for accurate behavior transfer across novel scenes.

\begin{table}[ht]
    \centering
    \caption{Ablation on scene context concatenation in BehaviorVAE using the VBD diffusion backbone. \textbf{With Context} improves spatial accuracy (mADE) and kinematic realism by enabling diverse, context-aware generation.}
    \label{tab:ablation_context}
    \resizebox{0.7\textwidth}{!}{
    \begin{tabular}{lccccc}
        \toprule
        \textbf{Setting} & \textbf{Meta} $\uparrow$ & \textbf{Kinematic} $\uparrow$ & \textbf{Interactive} $\uparrow$ & \textbf{Map} $\uparrow$ & \textbf{mADE} $\downarrow$ \\
        \midrule
        No Context & \textbf{0.7361} & 0.4481 & \textbf{0.7665} & \textbf{0.8615} & 1.0368 \\
        With Context & 0.7335 & \textbf{0.4511} & 0.7636 & 0.8563 & \textbf{1.0228} \\
        \bottomrule
    \end{tabular}
    }
\end{table}

\subsection{Ablation on Language Instruction Formats}
\label{subsec:prosim_instruction}

To determine the optimal language conditioning strategy, we evaluate three different formats of descriptions derived from the ProSim-Instruct-520k dataset:
\begin{itemize}
    \item \textbf{Text Only:} Utilizing the original auto-generated scene descriptions from ProSim-Instruct-520k~\cite{prosim} (e.g., \textit{"Make vehicle 1, vehicle 2... and the ego vehicle remain parked for the entire simulation"}).
    \item \textbf{Tag Only:} Employing simple, template-based sentences filled with discrete kinematic motion tags (e.g., \textit{"The target vehicle is turning left and accelerating"}).
    \item \textbf{Text + Tag:} A concatenation of both high-level scene descriptions and low-level motion tags.
\end{itemize}

As shown in Table~\ref{tab:ablation_language}, the \textbf{Tag Only} format achieves the lowest Minimum Average Displacement Error (mADE) of 1.4011 and the highest mADE reduction (15.60\%). This is expected, as explicit motion tags provide direct kinematic "shortcuts" that make trajectory matching easier for the model. However, we ultimately adopt the \textbf{Text + Tag} strategy for our framework. While it yields slightly higher mADE compared to the tag-only approach, it maintains the highest Map realism score (0.9024) by retaining crucial high-level scene context. This fosters a more diverse and comprehensive language-guided controllability rather than merely overfitting to kinematic templates. Furthermore, this observation highlights that future efforts in curating precise, manually annotated, and diverse language descriptions will be pivotal in unlocking the full potential of instruction-following in traffic simulation~\cite{langtraj}.

\begin{table}[ht]
    \centering
    \caption{Ablation on language instruction formats. The \textbf{Text + Tag} format balances low-level kinematic precision with high-level scene context.}
    \label{tab:ablation_language}
    \resizebox{0.8\textwidth}{!}{
    \begin{tabular}{lcccccc}
        \toprule
        \textbf{Format} & \textbf{Meta} $\uparrow$ & \textbf{Kinematic} $\uparrow$ & \textbf{Interactive} $\uparrow$ & \textbf{Map} $\uparrow$ & \textbf{mADE} $\downarrow$ & \textbf{mADE Gain} $\uparrow$ \\
        \midrule
        Uncond & 0.7665 & 0.4669 & 0.8030 & 0.8906 & 1.6601 & -- \\
        Text Only & 0.7685 & 0.4690 & 0.8050 & 0.8928 & 1.5790 & 4.88\% \\
        Tag Only & \textbf{0.7766} & \textbf{0.4832} & \textbf{0.8099} & 0.9015 & \textbf{1.4011} & \textbf{15.60\%} \\
        Text + Tag & 0.7742 & 0.4779 & 0.8062 & \textbf{0.9024} & 1.4655 & 11.72\% \\
        \bottomrule
    \end{tabular}
    }
\end{table}

\section{Discussion and Limitations}
\label{sec:app_limitations}

\begin{figure}[ht]
    \centering
    \includegraphics[width=0.8\linewidth]{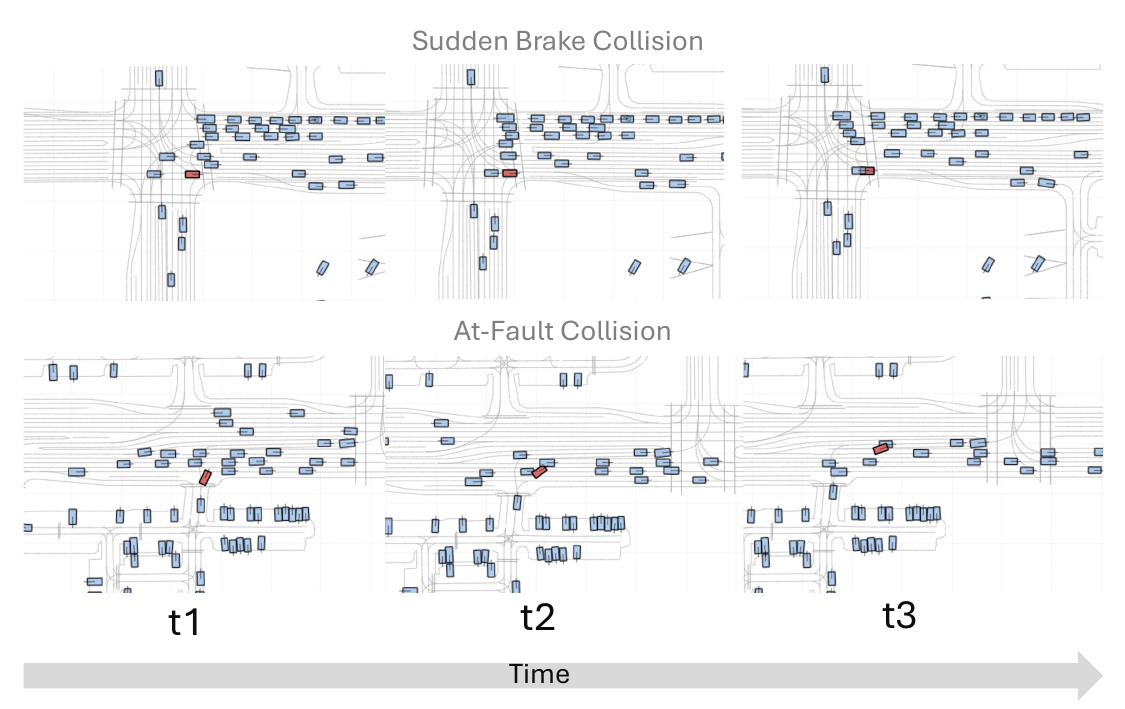}
    \caption{Failure cases during strong context domain shifts. \textbf{Top (Sudden Brake):} The agent strictly executes a stop instruction, ignoring trailing vehicles. \textbf{Bottom (At-Fault):} The agent forces a prompted maneuver in dense traffic, overriding reactive safety.}
    \label{fig:failurecase}
\end{figure}

While our modulation framework demonstrates effective controllable generation with data efficiency, challenges remain in handling out-of-distribution behaviors and instruction conflicts. In particular, retrieved counterfactual scenarios can be imbalanced: simple behaviors are common, while complex interactions such as overtaking or merging are harder to identify and label due to their rarity in the dataset. Furthermore, we notice that the model may prioritize control instructions over reactive safety, leading to failures such as \textit{sudden brake collisions} (e.g., ignoring trailing vehicles) or \textit{at-fault collisions} (e.g., forcing maneuvers in dense traffic), as illustrated in Fig.~\ref{fig:failurecase}. Future work can utilize reinforcement learning fine-tuning to mitigate these safety-related failures by better balancing instruction adherence with environment-aware reactions. Finally, our current experiments focus on transferring behaviors of a single controlled agent; future work will explore conditioning on multiple agents simultaneously to enable richer multi-agent interaction modeling.
 % Finally, our current experiments focus on transferring behaviors of a single controlled agent; future work will explore conditioning on multiple agents simultaneously to enable richer multi-agent interaction modeling.

% Finally, current synthetic metrics struggle to fairly weigh instruction-induced rule violations against baseline safety failures, underscoring the necessity for nuanced, context-aware evaluation protocols.

% ==========================================
% A. Experimental Setup and Metrics
% ==========================================
\section{Experimental Setup and Evaluation Metrics}
\label{sec:app_setup_metrics}

\subsection{Datasets and Oracle Validation Setup}

% In our evaluation in Sec 4.4, we feed ground-truth (GT) condition inputs (e.g., sketches, latent codes, or text instructions) directly to those agents with condition input within the scene, ensuring a direct measure of the model's trajectory synthesis performance.

In our evaluation for controllable simulation (Sec 4.2 and 4.3), we feed ground-truth (GT) condition inputs (e.g., sketches, latent codes, or text instructions) directly to the target agents within the scene, ensuring a direct measure of the model's trajectory synthesis performance. Conversely, for counterfactual and long-tail generation (Sec 4.4), we condition target agents on behavioral inputs transferred from context-matched agents to evaluate practical, non-oracle scenario synthesis.

\subsection{WOSAC Challenge Metrics Formulation}
The Waymo Open Sim Agent Challenge (WOSAC) evaluates simulation quality by computing negative log-likelihood (NLL) scores across nine statistical features, categorized into kinematics, agent interactions, and map adherence. The challenge requires simulating up to 128 agents for 8 seconds, generating 32 joint future samples per scenario.

For a given scenario $i$, target agent $a$, at timestep $t$, and statistical feature $F_j$, the likelihood of the ground-truth trajectory is given by:
\begin{equation}
    \text{NLL}(i, a, t, j) = -\log p_{i,j,a}(F_j(x^*(i, a, t)))
\end{equation}
where $p_{i,j,a}(\cdot)$ represents the empirical histogram distribution constructed from the simulated samples. A lower NLL indicates better alignment with real-world behavior.

The per-scenario metric is obtained by aggregating NLL values over valid timesteps:
\begin{equation}
    m(a, i, j) = \exp \left( -\frac{1}{N(i, a)} \sum_{t} v(i, a, t) \cdot \text{NLL}(i, a, t, j) \right)
\end{equation}
where $N(i, a) = \sum_{t} v(i, a, t)$ denotes the number of valid timesteps for agent $a$. The scenario-level metric is computed by averaging over all target agents $A_{\text{target}}$:
\begin{equation}
    m(i, j) = \frac{1}{|A_{\text{target}}|} \sum_{a \in A_{\text{target}}} m(a, i, j)
\end{equation}
Finally, the composite WOSAC Meta metric $M$ is a weighted average over all $F$ statistical features across $N$ scenarios:
\begin{equation}
    M = \frac{1}{N} \sum_{i=1}^{N} \sum_{j=1}^{F} w_j m(i, j), \quad \text{s.t. } \sum_{j=1}^{F} w_j = 1
\end{equation}

\subsubsection{Component Metric Definitions}
\begin{itemize}
    \item \textbf{Kinematic Metrics:}
    \begin{itemize}
        \item \textit{Linear Speed:} Magnitude of velocity, $\|v\| = \| \frac{x_{t+1}-x_t}{\Delta t} \|_2$.
        \item \textit{Linear Acceleration:} Magnitude of acceleration, $\| \frac{v_{t+1}-v_t}{\Delta t} \|_2$.
        \item \textit{Angular Speed:} Rate of heading change, $\omega = \frac{d(\theta_{t+1}, \theta_t)}{\Delta t}$.
        \item \textit{Angular Acceleration:} Rate of change of angular speed, $\frac{d(\omega_{t+1}, \omega_t)}{\Delta t}$.
    \end{itemize}
    \item \textbf{Interaction Metrics:}
    \begin{itemize}
        \item \textit{Distance to Nearest Object:} Signed distance calculated via the GJK algorithm.
        \item \textit{Collisions:} Boolean flag triggered when signed distance $< 0$.
        \item \textit{Time-to-Collision (TTC):} Estimated time until collision under constant velocity.
    \end{itemize}
    \item \textbf{Map Metrics:}
    \begin{itemize}
        \item \textit{Distance to Road Edge:} Signed distance to the nearest map boundary.
        \item \textit{Road Departure:} Boolean indicator of whether the agent exits the road boundary.
    \end{itemize}
    \item \textbf{Displacement Metrics:}
    \begin{itemize}
    
        \item \textit{Average Displacement Error (ADE):}
        Mean Euclidean distance between the predicted and ground-truth
        3D trajectories. For agent \(i\) in rollout \(r\) over valid
        timesteps \(T_i\), it is computed as
        \begin{equation}
            \mathrm{ADE}_{r,i}
            =
            \frac{1}{|T_i|}
            \sum_{t\in T_i}
            \left\|
            \mathbf{p}^{(r)}_{i,t}
            -
            \mathbf{p}^{\mathrm{log}}_{i,t}
            \right\|_2 .
        \end{equation}
        Averaging across all \(R\) rollouts and \(N\) agents gives
        \begin{equation}
            \mathrm{ADE}
            =
            \frac{1}{RN}
            \sum_{r=1}^{R}
            \sum_{i=1}^{N}
            \mathrm{ADE}_{r,i}.
        \end{equation}
    
        \item \textit{Minimum Average Displacement Error (mADE):}
        The minimum displacement error among all generated rollouts.
        It first averages \(\mathrm{ADE}_{r,i}\) over all \(N\) evaluated
        agents within each rollout and then selects the minimum across
        \(R\) rollouts:
        \begin{equation}
            \mathrm{mADE}
            =
            \min_{r\in\{1,\dots,R\}}
            \frac{1}{N}
            \sum_{i=1}^{N}
            \mathrm{ADE}_{r,i}.
        \end{equation}
    
    \end{itemize}
\end{itemize}

For further implementation details, refer to the WOSAC Challenge documentation~\cite{wosac}.

% PDMScore now matches the official implementation.
\subsection{PDMScore Formulation}

We adapt PDMScore~\cite{navsim} to evaluate controlled closed-loop rollouts on
the Waymo dataset. Compared with the original ego-centric formulation, our
adaptation (i) evaluates the controlled target agent, (ii) measures progress along
the queried condition trajectory when available and along the associated route
otherwise, and (iii) sets the traffic-light compliance term to one because
traffic-light states are not evaluated. The original score structure, component
weights, and remaining metric definitions are retained and implemented using
WOMD map and agent geometry.

The final score $S$ is the product of multiplicative compliance terms ($M$)
and weighted performance metrics ($W$):
\begin{equation}
\begin{aligned}
  S &= 100 \, M W, \\
  M &= NC \cdot DAC \cdot DDC \cdot TLC, \\
  W &= \frac{5EP + 5TTC + 2LK + 2HC}{14}.
\end{aligned}
\end{equation}
We set $TLC=1$ because traffic-light states are not evaluated in our setting.
The remaining metric components are defined as follows:

\begin{itemize}
  \item \textbf{No At-Fault Collision ($NC$):}
  A binary multiplier indicating whether the target agent avoids at-fault
  collisions. Collisions are detected using oriented bounding-box overlap, and
  at-fault status is classified from the relative positions and motions of the
  interacting agents.

  \item \textbf{Drivable-Area Compliance ($DAC$):}
  A binary multiplier indicating whether the target agent remains within the
  drivable region. We associate the rollout with nearby lane and drivable-area
  polylines and evaluate the distances of both the bounding-box center and
  corners. Brief off-road deviations are tolerated according to the thresholds
  specified in our implementation.

  \item \textbf{Driving-Direction Compliance ($DDC$):}
  A route-compliance multiplier that penalizes sustained motion opposite to the
  direction of the associated lane.

  \item \textbf{Ego Progress ($EP$):}
  Following the original PDMScore terminology, this metric measures the positive
  projected displacement of the target agent along the queried condition
  trajectory when available, or along the associated route otherwise. The
  resulting progress score is normalized and clipped to $[0,1]$.

  \item \textbf{Time-to-Collision Proxy ($TTC$):}
  A binary short-horizon safety score obtained by extrapolating the target agent
  under a constant-velocity assumption and checking for potential at-fault
  collisions within the TTC horizon.

  \item \textbf{Lane Keeping ($LK$):}
  A binary route-adherence score based on the smoothed lateral distance to the
  associated lane. We set $LK=0$ when this distance continuously exceeds the
  lane-keeping threshold for the configured duration.

  \item \textbf{History Comfort ($HC$):}
  A binary comfort score based on bounds on longitudinal and lateral
  acceleration, jerk, yaw rate, and yaw acceleration. Temporal smoothing and
  quantile-based aggregation reduce sensitivity to isolated noise in generated
  rollouts.
\end{itemize}

% These adaptations retain the structure of PDMScore while extending it to target-agent evaluation under queried counterfactual conditions.

\subsection{Additional Evaluation Metrics}
Beyond WOSAC and PDMScore, we report the following metrics in Sec. 4.4 of the main text to provide a comprehensive analysis of trajectory fidelity and generative diversity. For each scenario, metrics are computed exclusively on the target controlled agents across $K=32$ generated rollouts.

\begin{itemize}
    \item \textbf{Trajectory ADE (TrajADE):} Calculates the mean $L_2$ distance between the predicted trajectories and the unconditioned model rollout trajectories over the overlapping valid frames $T_{\text{overlap}}$. Averaged over the $K$ rollouts, it measures the overall spatial deviation:
    \begin{equation}
        \text{TrajADE} = \frac{1}{K} \sum_{k=1}^{K} \left( \frac{1}{|T_{\text{overlap}}|} \sum_{t \in T_{\text{overlap}}} \|x_{k,t}^{\text{pred}} - x_t^{\text{ref}} \|_2 \right)
    \end{equation}

    \item \textbf{Control ADE:} Measures the geometric fidelity between the synthesized trajectory and the queried condition trajectory after applying a local coordinate alignment (translation and rotation). This alignment removes ego-motion bias to strictly evaluate control adherence. Averaged over the $K$ rollouts:
    \begin{equation}
        \text{Control ADE} = \frac{1}{K} \sum_{k=1}^{K} \left( \frac{1}{|T_{\text{overlap}}|} \sum_{t \in T_{\text{overlap}}} \|x_{k,t}^{q, \text{local}} - x_{k,t}^{\text{pred, local}} \|_2 \right)
    \end{equation}

    \item \textbf{Success Rate (\%):} The average ratio of valid rollouts that successfully meet specific safety and goal completion criteria, averaged across all $N_{\text{cases}}$ scenarios:
    \begin{equation}
        \text{Success} = \frac{1}{N_{\text{cases}}} \sum_{i=1}^{N_{\text{cases}}} \left( \frac{n_{\text{success}, i}}{n_{\text{valid}, i}} \right) \times 100\%
    \end{equation}

    \item \textbf{Coverage:} Quantifies the exploration breadth of the generative rollouts. Following~\cite{bits}, we discretize the map into a set of grid cells $\mathcal{G}$, where each cell $g$ has a center $c_g$ and an area $\Delta A$. Let $d_g \in \{0, 1\}$ denote the drivable space mask ($1$ for drivable). We apply Kernel Density Estimation (KDE) over all rollouts and timesteps to obtain the spatial density $\hat{\rho}(x)$. A cell is defined as "covered" if its density exceeds a threshold $\tau$, denoted by the indicator function:
    \begin{equation}
        z_g = \mathbf{1}[\hat{\rho}(c_g) > \tau]
    \end{equation}
    From this, we define the Coverage Ratio ($C_{\text{all}}$) as follows:
    \begin{equation}
        C_{\text{all}} = \frac{1}{|\mathcal{G}|} \sum_{g \in \mathcal{G}} z_g 
    \end{equation}
\end{itemize}

% ==========================================
% B. 實作與架構細節
% ==========================================
\section{Implementation Details}
\label{sec:app_implementation}
\subsection{Inference and Decoding Strategy}
To explicitly balance control fidelity and natural driving priors, our synthetic scenario experiments utilize the SMART~\cite{smart} model trained on a 10\% subset of the full training data. During autoregressive decoding, we employ a logits-level Classifier-Free Guidance (CFG)~\cite{cfg} mechanism. At each decoding step, we extract unconditional logits ($L_{\text{base}}$) from the base model and conditional logits ($L_{\text{control}}$) from the control model. The final extrapolated logits are then computed as:
$$L_{\text{final}} = L_{\text{base}} + \omega \times (L_{\text{control}} - L_{\text{base}})$$
This mixed $L_{\text{final}}$ is subsequently converted to a probability distribution via Softmax, followed by standard Top-K and Top-P sampling strategies to predict the next trajectory token. To optimize the trade-off between instruction adherence and scene realism, we empirically configure the guidance scale $\omega$ based on the control modality: $\omega=0.5$ for Sketch, $\omega=0.7$ for Latent, and $\omega=1.0$ for Text conditioning.

\subsection{LoRA Base-Model Finetuning on the Full Training Set}
As a reference performance in Sec.~4.3 (Fig.~5), we additionally consider a setting where the control branch is trained jointly with LoRA adapters inserted into the base decoder using the full training dataset. This setting increases both adaptation capacity and data scale.

Concretely, we keep the pretrained SMART backbone frozen at the full-parameter level but inject LoRA adapters into the base decoder's attention projections 
% (\texttt{to\_q}, \texttt{to\_k}, \texttt{to\_v}, and \texttt{to\_out}) 
with a rank of 16, a scaling factor of 0.4, and a dropout of 0.05. The control branch remains active and uses the same FiLM-based conditioning interface as in the standard control experiments. In the sketch-conditioned example run, the control encoder utilizes a 16-step control horizon, 256-dimensional control tokens, 2 temporal transformer layers, 4 attention heads, and FiLM modulation at decoder blocks 0--5 with a FiLM hidden dimension of 256.

This setting is trained on the full Waymo training set (486,995 scenarios) using a batch size of 4, a learning rate of $5\times10^{-5}$, 16-bit mixed-precision training, and gradient clipping at 0.5 for up to 100 epochs.

\subsection{ProSim Unconditional Baseline Implementation}
\label{subsec:prosim_unconditional}

To obtain the true unconditional ProSim baseline, which has not been exposed to any paired control data during training, we mask all prompt inputs and retrain the model from scratch following the original ProSim training procedure~\cite{prosim}. We use the original optimizer and training hyperparameters and select the checkpoint after the validation loss has converged. The retrained model achieves an mADE of 2.641, comparable to 2.679 from the official ProSim checkpoint, providing a consistent unconditional baseline for evaluating our FiLM adaptation.

\subsection{BehaviorVAE Training Details}
Our BehaviorVAE was trained on 10,000 Waymo scenarios (approximately 2\% of the full training set), processing up to 64 agents per scene and predicting 41 future time steps. The model employs a 6-layer context encoder inherited from VBD~\cite{vbd}, which is jointly optimized with the VAE. Each agent's future trajectory is transformed into its local coordinate frame, represented as $(\Delta x, \Delta y, \sin \Delta \theta, \cos \Delta \theta)$, and processed by a 2-layer GRU (hidden size 512) followed by a single relative-geometry-aware self-attention layer to model inter-agent interactions. This extracts a 512-dimensional agent-specific latent variable without relying on a global scene latent. The posterior is parameterized by concatenating trajectory and context features, while the conditional prior is predicted solely from context features. The decoder, a 6-layer MLP, reconstructs trajectory deltas by conditioning on the latent code and a learned time embedding. The training objective minimizes a masked MSE reconstruction loss alongside a KL regularization term ($\mathcal{L} = \mathcal{L}_{\mathrm{recon}} + \beta \mathcal{L}_{\mathrm{KL}}$). Using latent sampling after a 1,000-step warm-up, the KL coefficient ($\beta$) is linearly annealed to 0.001 over the first 20,000 steps. The model is optimized using AdamW~\cite{adamw} (learning rate $3\times10^{-4}$, batch size 4, gradient clipping 1.0) in FP32 precision on two RTX 4080 GPUs for 200 epochs. At convergence, the training loss reached 0.0269 and the validation loss was 0.0389, establishing a diverse and structured behavior prior.

\subsection{Control Representation Encoders}
We adopt the same control-conditioning interface for both the SMART and VBD backbones, where each control modality is first encoded into a standardized 256-dimensional per-agent control token and injected into the frozen generative decoder through FiLM layers. For text control, both models utilize a frozen DistilBERT~\cite{sanh2019distilbert} (\texttt{distilbert-base-uncased}) backbone to process prompts (truncated to 384 tokens) from the ProSim-Instruct-520k corpus. We insert LoRA~\cite{lora} adapters (rank 16, scaling factor 0.4, dropout 0.05) into the attention projections of 6 adapted layers, and the resulting text features are linearly projected into the shared control space. For sketch control, each agent's relative $(x,y,\theta)$ trajectory over a 16-step control horizon is processed by a 2-layer temporal transformer (with 4 attention heads and a feed-forward dimension of 256) and pooled into the control token. Finally, for latent control, we retain the identical FiLM interface but replace the sequence encoders with a 3-layer MLP that maps the 512-dimensional BehaviorVAE latent codes directly into the 256-dimensional conditioning space. Across all settings, the backbone generative decoder remains completely frozen; only the modality-specific control branches and FiLM layers are optimized.

\subsection{FiLM Parameters and Structure}
The FiLM~\cite{film} module in both the SMART and VBD backbones utilizes direct feature-wise affine modulation:
\begin{equation}
    \mathrm{FiLM}(x,c) = \gamma(c) \odot x + \beta(c)
\end{equation}
where $x$ denotes the decoder hidden feature and $c$ denotes the control token. In both models, the FiLM MLP takes a 256-dimensional conditioning feature as input and predicts channel-wise modulation parameters for the decoder hidden state.

For the SMART backbone, FiLM is attached to decoder blocks 0--5, with a FiLM hidden dimension of 256. For VBD, FiLM is attached to denoiser blocks 0--1, with a FiLM hidden dimension of 512 in the referenced text-conditioned run. Crucially, in both implementations, FiLM is initialized close to the identity transformation. The final FiLM layer is initialized with very small weights, the scale branch is biased toward $\gamma \approx 1$, and the shift branch is biased toward $\beta \approx 0$ (with a minor random perturbation added to avoid exact identity at initialization). This design stabilizes control fine-tuning by ensuring the pretrained decoder behavior is perfectly preserved at the beginning of training, allowing the control branch to gradually learn non-trivial modulations.

\subsection{Optimization Hyperparameters}
For control fine-tuning, both models are optimized using AdamW under 16-bit mixed-precision training. Unless otherwise noted, the base decoder remains frozen, and only the control modules are updated. For the SMART backbone, the learning rate is set to $5\times10^{-5}$ with no warm-up, a minimum learning-rate ratio of 0.05, gradient clipping at 0.5, and synchronized batch normalization. For VBD control training, we use a learning rate of $1\times10^{-4}$ for the Control branch, with a weight decay of 0.01 and gradient clipping via value clamping at 5.0. 

Both SMART and VBD are trained under receding-horizon closed-loop control. In VBD, replanning is performed every 11 steps, utilizing a prediction horizon of 40 steps and 5 denoising steps per planning update; additionally, during closed-loop training, we generate 8 candidate trajectories and select the one closest to the ground truth. In SMART, control is similarly applied in a receding-horizon manner, processing 11 historical steps and 80 future steps in the decoder, with control signals injected over a 16-step control horizon (window-aligned latent indexing is used for latent control).

\subsection{Context Matching Heuristic Filter}
To retrieve compatible donor scenarios, we use a non-leaky context-matching heuristic based only on observed history. For each query agent, candidate agents are represented by a SMART history embedding and five historical descriptors: \textbf{(i) Dynamic}, including speed, acceleration, yaw rate, curvature, and displacement; \textbf{(ii) Trajectory}, the recent ego-aligned trajectory; \textbf{(iii) Map Context}, the local map composition; \textbf{(iv) Map kNN}, nearby map geometry and types; and \textbf{(v) Neighbor Context \& kNN}, surrounding traffic layout, relative motion, and agent types.

Retrieval proceeds in three stages. \textbf{First}, candidates are restricted to the same agent type and the query scenario is excluded. \textbf{Second}, a coarse pool is formed by combining top candidates ranked by encoder-feature similarity and trajectory distance. \textbf{Third}, candidates are re-ranked using
\begin{equation}
\begin{aligned}
s ={} & w_{\mathrm{feat}} s_{\mathrm{feat}} + 
        w_{\mathrm{traj}} d_{\mathrm{traj}} + 
        w_{\mathrm{map\_knn}} d_{\mathrm{map\_knn}} + 
        w_{\mathrm{nbr\_knn}} d_{\mathrm{nbr\_knn}} \\ % <-- 修正 1: 改成 \\ 換行
      & - w_{\mathrm{dyn}} d_{\mathrm{dyn}} - 
        w_{\mathrm{map}} d_{\mathrm{map}} - 
        w_{\mathrm{nbr}} d_{\mathrm{nbr}},
\end{aligned}
\end{equation}
which rewards feature similarity and penalizes descriptor distances. % <-- 修正 2: 上面的 \mathrm{...} 內底線皆改為 \_

We use geometry-aware weights of 
$w_{\mathrm{feat}}=0.18$, 
$w_{\mathrm{traj}}=0.40$, 
$w_{\mathrm{map\_knn}}=0.22$, 
$w_{\mathrm{nbr\_knn}}=0.20$, 
$w_{\mathrm{dyn}}=0.08$, 
$w_{\mathrm{map}}=0.04$, and 
$w_{\mathrm{nbr}}=0.03$. 
We further require $d_{\mathrm{traj}} \le 0.08$, $d_{\mathrm{map\_knn}} \le 1.4$, and $d_{\mathrm{nbr\_knn}} \le 1.3$. If no candidate passes these thresholds, a score-based fallback retrieves the minimum required number of matches.

\subsection{Qualitative Examples of Language Prompts}
\label{subsec:prompt_examples}

We show five prompts from scenario \texttt{b6975ecc8d04afde} to illustrate our \textbf{Text + Tag} format, which combines scene descriptions with motion tags.

\begin{tcolorbox}[
    colback=gray!5!white,      % 內部背景顏色 (極淺的灰色)
    colframe=black!70,         % 邊框顏色 (深灰色)
    boxrule=1pt,               % 邊框粗細
    arc=3pt,                   % 圓角半徑
    title={\textbf{Scenario ID:} \texttt{b6975ecc8d04afde}},
    fonttitle=\bfseries,       % 標題字體加粗
    coltitle=white,            % 標題字體顏色
    bottom=3mm, top=3mm, left=3mm, right=3mm % 內部邊距
]

\textbf{Agent 777:} Meanwhile, vehicle 1, the target vehicle, and vehicle 2 move in a straight line throughout the scenario. The target vehicle is moving straight between 0.0 seconds and 9.0 seconds and decelerating between 5.0 seconds and 6.0 seconds.

\textbf{Agent 781:} The target vehicle makes a left turn from 4.0 seconds onwards. The target vehicle is turning left between 1.0 seconds and 6.0 seconds, slowing down to a stop between 5.5 seconds and 9.0 seconds, and moving straight between 6.0 seconds and 8.5 seconds.

\textbf{Agent 877:} At the start of the simulation, have the target vehicle accelerate and then make a right turn. The target vehicle first accelerates, then makes a right turn, and eventually stops. The target vehicle is accelerating between 0.0 seconds and 1.0 seconds and turning right between 0.0 seconds and 1.5 seconds.

\textbf{Agent 773:} The target vehicle makes a left turn from the start to the end of the scenario. The target vehicle is turning left between 0.0 seconds and 9.0 seconds and keeping a steady speed between 5.5 seconds and 6.5 seconds.

\textbf{Agent 768:} A cluster of vehicles, including the target vehicle, vehicle 1, vehicle 2, and others, remain parked throughout the simulation. The target vehicle is parked between 0.0 seconds and 9.0 seconds.

\end{tcolorbox}